\documentclass{article}
\AtBeginDocument{%
  }

\usepackage{algorithm}
\usepackage{algpseudocode}
\usepackage{amsmath}
\usepackage{amsfonts}
\usepackage{graphicx}
\usepackage{booktabs}
\usepackage{tabularx}
\usepackage[margin=1in]{geometry}
\usepackage{adjustbox}
\usepackage{threeparttable}

\pdfoutput=1
\usepackage{times}
\usepackage{latexsym}

\usepackage[T1]{fontenc}
\usepackage[utf8]{inputenc}
\usepackage{tikz}
\usepackage{forest}
\forestset{
  forked edges/.style={for tree={edge path={\noexpand\path[\forestoption{edge}](!u.parent anchor)--+(5pt,0)|-.child anchor)\forestoption{edge label};}},},
  folder/.style={grow'=0},
  fork sep/.style={s sep=#1},
}
\usepackage{microtype}
\usepackage{inconsolata}
\usepackage[table,xcdraw]{xcolor}
\usepackage{tcolorbox}
\usepackage{hyperref}

\definecolor{child-blue}{RGB}{40, 116, 166}
\definecolor{grandchild-blue}{RGB}{52, 152, 21}
\definecolor{greatgrandchild-blue}{RGB}{133, 193, 233}
\definecolor{leaf-blue}{RGB}{214, 234, 248}
\definecolor{child-green}{RGB}{20, 143, 119}
\definecolor{child-yellow}{RGB}{183, 149, 11}
\definecolor{child-brown}{RGB}{211, 84, 0}
\definecolor{child-purple}{RGB}{136, 78, 160}
\definecolor{EntitiesColor}{HTML}{6A8ACF}
\definecolor{EventsColor}{HTML}{D67C7C}
\definecolor{hidden-draw}{HTML}{FFFFFF}

\forestset{
    root style/.style={draw, thick, rounded corners, fill=gray!30},
    child style/.style={draw, thick,rounded corners, fill=blue!20},
    grandchild style/.style={draw, thick, rounded corners, fill=green!20},
    greatgrandchild style/.style={draw, thick, rounded corners, fill=yellow!20},
    leaf style/.style={draw, thick, rounded corners, fill=red!20},
    conceptsleaf style/.style={draw=child-purple, thick, rounded corners, fill=child-purple!15},
}

\tikzset{
  my-box/.style={
    rectangle,
    draw=hidden-draw,
    rounded corners,
    text opacity=1,
    minimum height=1.5em,
    minimum width=40em,
    inner sep=2pt,
    align=center,
    line width=0.8pt,
  },
  leaf/.style={
    my-box,
    minimum height=1.5em,
    text=black,
    align=center,
    font=\normalsize,
    inner xsep=2pt,
    inner ysep=4pt,
    line width=0.8pt,
  }
}

\begin{document}

\title{Small Language Models (SLMs) Can Still Pack a Punch: A Survey}

\author{Akanksha Gupta, Bijo Thomas, Harshita Asnani, \\
Phanindra Reddy Madduru, Samia Feroze, \\
Shreyas Subramanian, Vikram Elango, Mecit Gungor\\
\texttt{\{akankshg, bijothom, asnaharn, maddurup, samfero, subshrey, evikram, mecit\}@amazon.com}}

\maketitle

\begin{abstract}
As foundation AI models continue to increase in size, an important question arises - is massive scale the only path forward? This survey of about 160 papers presents a family of Small Language Models (SLMs) in the 1 to 9 billion parameter range that demonstrate smaller models can perform as well, or even outperform large models. We explore task agnostic, general purpose SLMs, task-specific SLMs and techniques to create SLMs that can guide the community to build models while balancing performance, efficiency, scalability and cost. Furthermore we define and characterize SLMs' effective sizes, representing increased capability with respect to LLMs.
\end{abstract}

\section{Introduction}
\label{introduction}

Large Language Models (LLMs) refer to Transformer-based language models (from \cite{vaswani}) with billions of parameters, which exhibit surprising abilities not present in smaller models. LLMs have had far reaching impact on academic research related to Language modeling as well as industry adoption. Several papers and surveys cover traditional LLMs - for example \cite{survey1} by Zhao \emph{et~al.} provides a comprehensive review of recent advances in LLMs.  The paper discusses key techniques for developing LLMs, including scaling laws, emergent abilities, distributed training algorithms, eliciting abilities through prompting, and aligning models to human values. The review also covers recent progress in pre-training, adaptation, utilization, and capability evaluation of LLMs. Other recent surveys on LLMs such as \cite{survey2} also cover similar topics, but additionally explores practical applications, productivity tools, prompting techniques, limitations and future challenges. Surveys such as \cite{survey2, survey3, survey4} all generally cover models that have more than 10B parameters, referred to as Large or Foundational models with a cursory mention of smaller models for language modeling.  Independently, there has been a growing interest in smaller language models. To the best of our knowledge, this paper presents a unique view on SLMs released recently which perform as well, or sometimes outperform larger counterparts. In summary, the main contributions are as follows:

\begin{itemize}
\itemsep0em 
\item We present an in-depth analysis of recent advancements in SLMs, highlighting specific SLMs with  their design, architecture, and the innovative techniques that enable them to achieve performance comparable to, or in some cases, surpassing that of larger models.

\item We categorize various SLMs based on their size, application domains, performance and training techniques, providing a comprehensive overview of the current SLM landscape and illustrating how these models can be effectively utilized in resource-constrained environments. Figure 1 provides a mind map of different ways to categorize SLMs.

\item We surface performance comparisons of SLMs with traditional LLMs and highlight the capacity and effective model sizes with respect to this performance. 

\end{itemize}

Since there is no \textit{standard} definition for SLMs that has surfaced so far, we make the following two clarifications: 1. A universally agreed upon line distinguishing SLMs vs LLMs cannot be drawn. As such we cover several SLMs that are in the few billion range, but see clusters of models in the 1B, 7B and 13B parameters; and 2. while thousands of narrow models have been created in the NLP and vision space, they differ from SLMs where a basic level of reasoning and language understanding is required to achieve good performance at multiple, or a single task. 

We note that state-of-the-art techniques allow 8 bit Adam training for 7B parameter models on a single consumer grade NVIDIA RTX 4090 GPUs with 24 GB of memory \cite{galore}. Additionally, 7B models like llama and Mistral are widely provided as an option commercially by LLM API providers like Amazon and Microsoft. As such our definition of SLMs includes general purpose language models with less than 8B parameters. In cases where the authors of papers with models up to 13B parameters, we include these as exceptions.

In the sections that follow, we begin with describing different types of 
SLMs, including task-agnostic, task-specific models and follow up with 
approaches to create SLMs. Section~\ref{sec:methodology} describes our 
survey methodology and the treatment of reported results.

\subsection{Survey Methodology}
\label{sec:methodology}

We identified papers through Google Scholar, Semantic Scholar, and 
arXiv (cs.CL, cs.LG, cs.AI) using queries related to small language 
models, model compression, knowledge distillation, and on-device 
inference. We also performed citation tracking from widely adopted 
models and monitored the Hugging Face Open LLM Leaderboard for newly 
released models in the target parameter range. The majority of papers 
we cover were published between 2023 and 2025. We report benchmark 
results as published in the original papers or as listed on the Open 
LLM Leaderboard; we have not independently replicated any reported 
results, and evaluation conditions may vary across studies.

\begin{figure}[h!]

\hspace{-1cm}
\fbox{\includegraphics[width=1.15\textwidth, trim={0.1cm 0.1cm 0.1cm 0.1cm},clip]{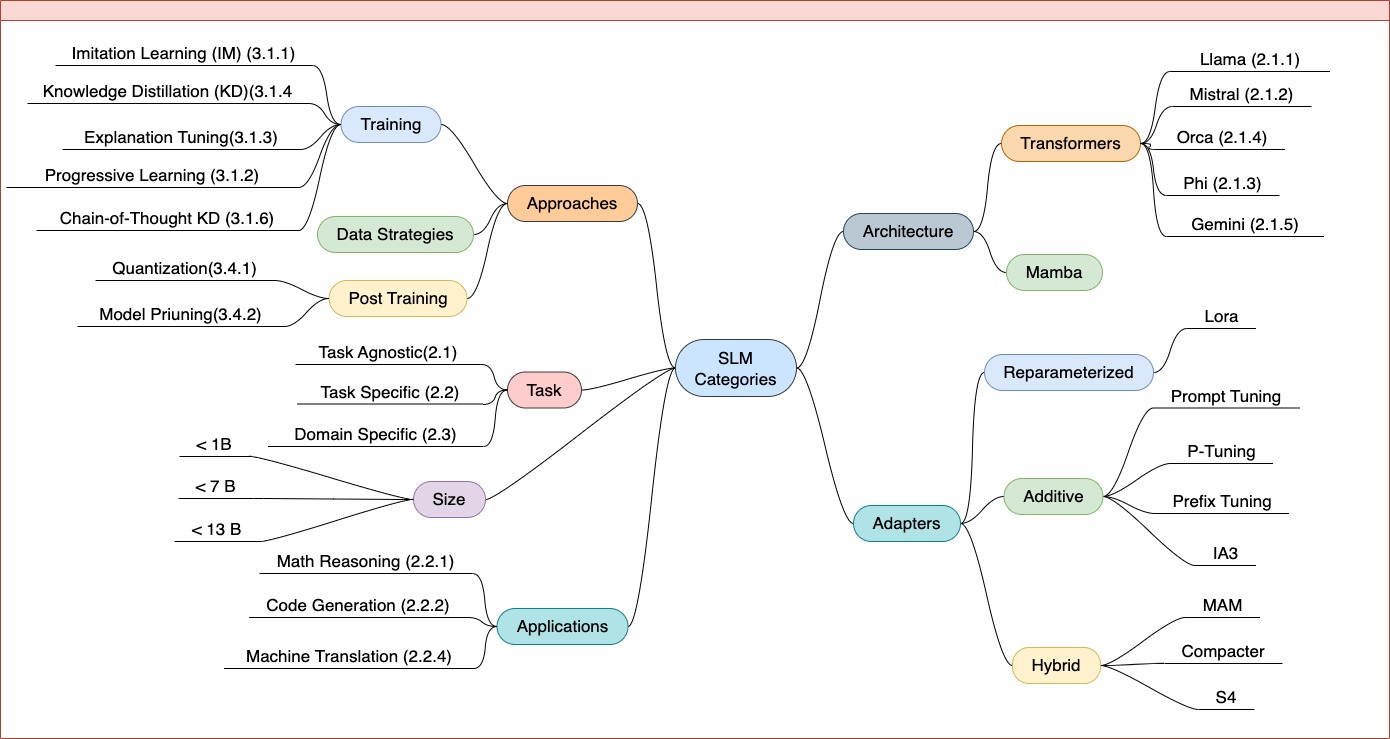}}
\caption{Mind map of topics covered in the paper}
\label{fig:organization}
\end{figure}

\section{Types of SLMs}

\subsection{Task Agnostic SLMs}
In the NLP field, researchers have explored training small, task-specific models with excellent performance in specific tasks but limited general language abilities. This contrasts with larger models, which develop broader, task-agnostic skills, including some reasoning and understanding. The extent to which small models can achieve this remains uncertain. For instance, the TinyStories model (10M parameters\footnote{https://arxiv.org/abs/2305.07759}) successfully generated coherent English stories using a synthetic dataset created by larger models (GPT 3.5 and GPT 4). However, as indicated in the preliminary Table \ref{tab:1} of Appendix \ref{app:1} exploring SLMs responses show that several billion parameters at minimum are required to generate good responses. Presently, models with several billion parameters are considered effective in task-agnostic settings, but more standardized, comprehensive benchmarks across SLMs are needed to fairly evaluate them. The next few sections will explore some successful SLMs that rival their larger counterparts; architectural parameters of these models are shown in Table \ref{tab:2} for comparison.

\begin{table}[h!]
\centering
\label{tab:2}
\caption{Comparison of task agnostic SLM Model Parameters where available}

\hspace{-0.85in}
\begin{tabular}{|l|p{1cm}p{1cm}p{1cm}p{1cm}p{1cm}p{1cm}p{1cm}p{1cm}p{1cm}|}
\hline
\textbf{Model Name} & \textbf{dim} & \textbf{layers} & \textbf{head dim} & \textbf{hidden dim} & \textbf{heads} & \textbf{kv heads} & \textbf{window} & \textbf{context len} & \textbf{vocab size} \\ 
\hline
Llama2 7B & 4096 & 32 & - & 11008 & 32 & - & - & 4096 & 32000 \\
TinyLlama 1.1B & 2048 & 22 & - & 5632 & 16 & - & - & 2048 & 32000 \\
Mistral 7B & 4096 & 32 & 128 & 14336 & 32 & 8 & 4096 & 8192 & 32000 \\
Phi 3B Mini & 3072 & 32 & 32 & - & - & 8 & - & 8192 & 32000 \\
Phi-1.5B & - & 24 & 64 & - & 32 & - & - & 2048 & - \\
Phi-1 & - & 24 & 64 & 2048 & 32 & - & - & 2048 & - \\
Chuxin 1.8B & - & 24 & - & 5632 & 32 & - & - & 1000000 & 102400 \\
Phi-1-Small & - & 20 & 64 & 1024 & 16 & - & - & 2048 & - \\
Gemini Nano 1\&2 & - & - & - & - & - & - & - & - & - \\
Gemma 2B & 2048 & 18 & 256 & 32K & 8 & 1 & - & 8192 & 256128 \\
Stable LM 1.6B & - & - & 24 & 2048 & 32 & - & - & 4096 & 100352 \\
\hline
\end{tabular}

\end{table}

\subsubsection{\textbf{Qwen:}} 

Qwen series (Alibaba Group) trained on 3 trillion tokens demonstrated that smaller models (1.8B-14B parameters) can achieve comparable or superior performance through optimized design. Qwen architecture implements enhanced tokenization with byte pair encoding (BPE) and approximately 152K vocabulary, untied embedding, RoPE (Rotary Positional Embedding) with FP32 precision for the inverse frequency matrix, and strategic bias in the QKV attention layer. The model employs pre-normalization with RMSNorm and SwiGLU activation functions, with feed-forward network dimension reduced to 8/3 of hidden size. QWEN-14B outperforms previous 13B SOTA models in general language tasks, while CODE-QWEN demonstrates superior code generation performance. MATH-QWEN-7B-CHAT surpasses Minerva-8B, approaching Minerva-62B and GPT-3.5 capabilities. Flash Attention and AdamW optimizer with tuned hyperparameters further enhance efficiency.

Qwen2\footnote{https://arxiv.org/abs/2407.10671} expanded across 0.5B-72B parameters with Grouped Query Attention (GQA) and Dual Chunk Attention (DCA) with YARN. Benchmark results: Qwen2-1.5B achieved 17.24\% on MMLU pro versus previous 9.80\%, while Qwen2-0.5B demonstrated 37.9\% on MMLU, 29.9\% on HumanEval, and 40.1\% on GSM8K. In coding, Qwen2-7B achieved 79.9\% on HumanEval and 67.2\% on MBPP. Qwen2-57B-A14B matched 30B dense models while activating only 14B parameters. Both models implement Flash Attention and optimized AdamW, but Qwen2 introduces fine-grained experts in MoE architecture and enhanced multilingual capabilities. Long-context capabilities through Needle in a Haystack showed smaller models effectively process up to 32K tokens, with larger variants handling 128K tokens.

Qwen3 dense models achieve performance parity: Qwen3-1.7B/4B/8B/14B match Qwen2.5-3B/7B/14B/32B respectively, while Qwen3-4B rivals Qwen2.5-72B-Instruct (as per Qwen3 technical report). This results from training on 36 trillion tokens (double Qwen2.5's 18 trillion) across 119 languages, with enhanced STEM, coding, and reasoning data. Models incorporate hybrid thinking modes, switching between rapid response and step-by-step reasoning. This was implemented through four-stage post-training: long chain-of-thought cold start, reasoning-based reinforcement learning, thinking mode fusion with instruction-tuning data, and general RL across 20+ domain tasks.

\begin{figure}[h!]
\includegraphics[width=\linewidth, trim=0 0 0 1cm, clip]{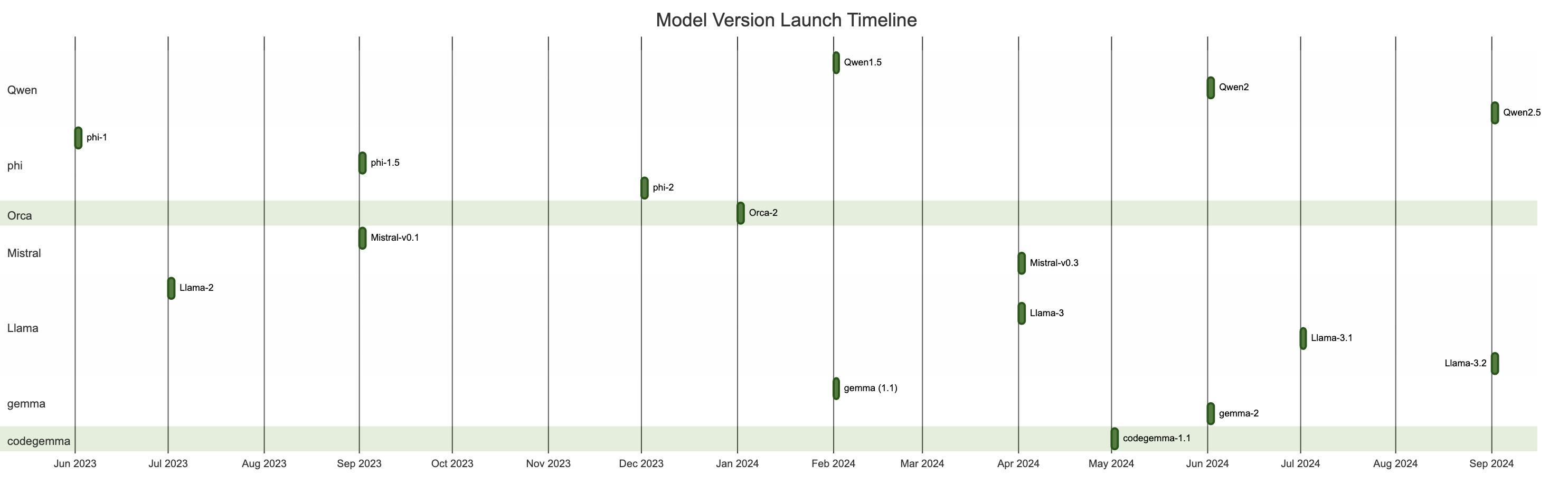}
\vspace{-0.7cm}
\caption{SLM model launch timeline}
\vspace{-0.1cm}
\label{fig:eqsize}
\end{figure}

\subsubsection{\textbf{Llama family:}}
The Llama2 paper included 7B, 13B, 34B and 70B variants; all pretrained, then converted to chat models through several months of SFT and aligned using RLHF on Meta's Research Super Cluster with NVIDA A100 GPUs.\footnote{https://arxiv.org/abs/2307.09288} Llama3 \footnote{https://arxiv.org/abs/2404.14047v3}, a low-bit quantized model trained on 15T tokens shows reduced memory and computational requirements with impressive performance. The study emphasizes high-quality datasets' importance, showcasing robustness for various quantization methods, even in ultra-low bit-width scenarios. Llama 2 architecture has been used to build state-of-the-art SLMs like Phi 3 models (Section \ref{sec:phi}) and Chuxin 1.8B; Chuxin has 1M context length, trained with 2.3 Trillion tokens. After 2 epochs, Chuxin's results are competitive with Gemma and Qwen on ARC, BoolQ, Hellaswag, SciQ, PiQA and Winogrande (from Qwen, chuxin and gemma technical reports).

The Llama3 herd \footnote{https://arxiv.org/abs/2407.21783} (2024) comprises models with \textbf{8B}, 70B and 405B parameters and 128,000 token context window. These models are trained on 15 Trillion tokens (vs. 1.8T for Llama2), enhancing coding, reasoning, and tool usage capabilities. Llama 8B beats Gemma 9B and Mistral 7B on benchmarks including MMLU, IFEval, HumanEval, GSM8K, MATH, BFCL, Nexus and MGSM. Being multi-modal, Llama 3 integrates image, video, and speech processing through a compositional approach, achieving competitive recognition results. Models are publicly available with Llama Guard 3 for safety.

Llama 3.2 series includes 1B and 3B parameter models supporting 128,000 token context length. These models are optimized for on-device applications (summarization, instruction following, content rewriting) on mobile and edge devices. They are compatible with Qualcomm and MediaTek hardware, fine-tuned for Arm processors. Despite smaller size, the 1B and 3B models demonstrate state-of-the-art performance within their class. Quantized versions provide reduced memory footprints and faster on-device inference without compromising accuracy. Llama 3.2 models are available on llama.com and Hugging Face, supported by AMD, AWS, Databricks, Dell, Google Cloud, Groq, IBM, Intel, Microsoft Azure, NVIDIA, Oracle Cloud, and Snowflake.

TinyLlama, a 1.1B parameter model using Llama architecture, trained on 1 trillion tokens for 3 epochs using Slimpajama and StarCoder datasets \cite{Li2023StarCoderMT}. TinyLlama incorporates Rotary Positional Embedding (RoPE) for positional information\cite{su2023roformer}, similar to PaLM, Llama, and Qwen \cite{chowdhery2023palm,touvron2023llama}. RMSNorm is applied for input normalization \cite{zhang2019root}. The model adopts SwiGLU, combining Swish and Gated Linear Unit, similar to Llama2.\cite{misra2019mish} TinyLlama employs grouped-query attention (GQA), dividing key-value heads into groups. Speed optimizations include Fully Sharded Data Parallel (FSDP) for efficient multi-GPU training and Flash Attention for improved attention mechanisms. This enables 24,000 tokens per second per A100-40G GPU and fits 1.1B model within 40GB GPU RAM. Compared to Pythia-1.0B and MPT-1.3B, TinyLlama shows superior training speed, requiring significantly fewer GPU hours \footnote{https://arxiv.org/abs/2401.02385}. TinyLLaVA \footnote{https://arxiv.org/abs/2402.14289} explores small-scale Large Multimodal Models (LMMs) with various vision encoders and connection modules. TinyLLaVA-3.1B outperforms existing 7B models like LLaVA-1.5 and Qwen-VL.

Except for WinonGrande \cite{Sakaguchi2019AnAW}, TinyLlama outperforms models on HellaSwag, OpenBookQA, ARC, BoolQ and PIQA \cite{Zellers2019HellaSwagCA,Mihaylov2018CanAS, Bisk2019PIQARA}. TinyLlama approaches or beats Llama2's performance for BoolQ, MMLU and PIQA while lagging on other benchmarks \cite{Hendrycks2020MeasuringMM,Zheng2023CodeGeeXAP}.

\begin{table}
\caption{Benchmark Results}
\label{tab:benchmark}
\small
\begin{threeparttable}
\begin{adjustbox}{width=\textwidth}
\begin{tabular}{|l|l|l|l|l|l|l|l|l|l|}
\toprule
Size & Type & Model & Average & IFEVal & BBH & MATH & GPQA & MUSR & MMLU-Pro \\
\midrule
0.5B & Base & Qwen1.5-0.5B & 5.35 & 17.06 & 5.04 & 1.74 & 0.56 & 4.30 & 3.41 \\
0.5B & Instruct & Qwen1.5-0.5B-Chat & 5.68 & 18.07 & 4.32 & 0.68 & 2.57 & 6.06 & 2.36 \\
0.5B & Base & Qwen2-0.5B & 7.22 & 18.73 & 7.92 & 2.64 & 1.45 & 4.60 & 8.00 \\
0.5B & Instruct & Qwen2-0.5B-Instruct & 6.59 & 22.47 & 5.88 & 2.87 & 0.00 & 2.41 & 5.90 \\
0.5B & Base & Qwen2.5-0.5B & 6.55 & 16.27 & 6.95 & 3.93 & 0.00 & 2.08 & 10.06 \\
0.5B & Instruct & Qwen2.5-0.5B-Instruct & 10.11 & 31.53 & 8.17 & 10.35 & 1.23 & 1.37 & 8.00 \\
0.6B & Base & Qwen3-0.6B\tnote{$\dagger$} & -- & -- & 41.47 & 32.44 & 26.77 & -- & 24.74 \\

1.3B & Base & phi-1 & 5.57 & 20.68 & 4.27 & 0.98 & 2.01 & 3.70 & 1.80 \\
1.5B & Base & DeepSeek-R1-Distill-Qwen-1.5B & 10.35 & 34.63 & 4.73 & 16.92 & 0.78 & 2.97 & 2.08 \\
1.5B & Base & Qwen2-1.5B & 10.45 & 21.13 & 11.78 & 7.02 & 1.90 & 3.59 & 17.24 \\
1.5B & Instruct & Qwen2-1.5B-Instruct & 14.14 & 33.71 & 13.70 & 7.18 & 1.57 & 12.03 & 16.68 \\
1.5B & Base & Qwen2.5-1.5B & 13.85 & 26.74 & 16.66 & 9.14 & 4.70 & 5.27 & 20.61 \\
1.5B & Instruct & Qwen2.5-1.5B-Instruct & 18.43 & 44.76 & 19.81 & 22.05 & 0.78 & 3.19 & 19.99 \\
1.7B & Base & Qwen3-1.7B\tnote{$\dagger$} & -- & -- & 54.47 & 43.50 & 28.28 & -- & 36.76 \\

1.8B & Base & Qwen1.5-1.8B & 9.27 & 21.54 & 9.76 & 3.17 & 7.38 & 3.96 & 9.80 \\
1.8B & Instruct & Qwen1.5-1.8B-Chat & 9.26 & 20.19 & 5.91 & 1.96 & 6.38 & 12.18 & 8.93 \\
1B & Base & Llama-3.2-1B & 4.20 & 14.78 & 4.37 & 1.21 & 0.00 & 2.56 & 2.26 \\
1B & Instruct & Llama-3.2-1B-Instruct & 14.44 & 56.98 & 8.74 & 7.02 & 3.36 & 2.97 & 7.58 \\
2.7B & Base & phi-2 & 15.53 & 27.39 & 28.04 & 2.95 & 2.91 & 13.84 & 18.09 \\
2B & Base & codegemma-1.1-2b & 7.13 & 22.94 & 7.55 & 1.28 & 2.01 & 5.93 & 3.09 \\
2B & Instruct & gemma-1.1-2b-it & 8.05 & 30.67 & 5.86 & 1.81 & 2.57 & 2.02 & 5.37 \\
2B & Base & gemma-2-2b & 10.13 & 19.93 & 11.76 & 2.87 & 1.68 & 11.43 & 13.11 \\
2B & Instruct & gemma-2-2b-it & 17.05 & 56.68 & 17.98 & 0.08 & 3.24 & 7.08 & 17.22 \\
2B & Instruct & gemma-2b-it & 7.49 & 26.90 & 5.21 & 2.04 & 3.80 & 3.03 & 3.92 \\
3.8B & Instruct & Phi-3-mini-128k-instruct & 26.34 & 59.76 & 37.10 & 14.05 & 9.06 & 7.71 & 30.38 \\
3.8B & Instruct & Phi-3-mini-4k-instruct & 25.97 & 56.13 & 39.27 & 11.63 & 9.28 & 7.64 & 31.85 \\
3.8B & Instruct & Phi-3.5-mini-instruct & 28.18 & 57.75 & 36.75 & 19.64 & 11.97 & 10.10 & 32.91 \\
3.8B & Instruct & Phi-4-mini-instruct & 29.41 & 73.78 & 38.74 & 16.99 & 7.94 & 6.45 & 32.58 \\
3B & Base & Llama-3.2-3B & 8.70 & 13.37 & 14.23 & 1.89 & 2.35 & 3.81 & 16.53 \\
3B & Instruct & Llama-3.2-3B-Instruct & 24.20 & 73.93 & 24.06 & 17.67 & 3.80 & 1.37 & 24.39 \\
3B & Base & Qwen2.5-3B & 18.10 & 26.90 & 24.30 & 14.80 & 6.38 & 11.76 & 24.48 \\
3B & Instruct & Qwen2.5-3B-Instruct & 27.16 & 64.75 & 25.80 & 36.78 & 3.02 & 7.57 & 25.05 \\
4B & Base & Qwen1.5-4B & 11.77 & 24.45 & 16.25 & 5.29 & 3.58 & 4.82 & 16.22 \\
4B & Instruct & Qwen1.5-4B-Chat & 12.63 & 31.57 & 16.30 & 2.79 & 2.24 & 7.36 & 15.51 \\
4B & Base & Qwen3-4B\tnote{$\dagger$} & -- & -- & 72.59 & 54.10 & 36.87 & -- & 50.58 \\
7B & Base & Llama-2-7b-hf & 8.81 & 25.19 & 10.35 & 1.74 & 2.24 & 3.76 & 9.57 \\
7B & Instruct & Llama-2-7b-chat-hf & 9.61 & 39.86 & 4.46 & 1.96 & 0.45 & 3.28 & 7.64 \\
7B & Instruct & Phi-3-small-128k-instruct & 31.97 & 63.68 & 45.63 & 20.26 & 8.95 & 14.50 & 38.78 \\
7B & Instruct & Phi-3-small-8k-instruct & 32.34 & 64.97 & 46.21 & 18.87 & 8.28 & 16.77 & 38.96 \\
7B & Base & Orca-2-7b & 14.40 & 21.83 & 22.43 & 1.96 & 1.45 & 24.09 & 14.65 \\
7B & Instruct & gemma-1.1-7b-it & 17.69 & 50.39 & 15.93 & 4.91 & 5.82 & 11.51 & 17.60 \\
7B & Base & gemma-7b & 15.44 & 26.59 & 21.12 & 7.40 & 4.92 & 10.98 & 21.64 \\
7B & Instruct & gemma-7b-it & 13.07 & 38.68 & 11.94 & 2.95 & 4.59 & 12.53 & 7.72 \\
7B & Base & DeepSeek-R1-Distill-Qwen-7B & 14.99 & 40.38 & 7.88 & 19.56 & 3.91 & 3.55 & 14.68 \\
7B & Base & Qwen1.5-7B & 16.02 & 26.84 & 23.08 & 9.29 & 6.49 & 9.16 & 21.29 \\
7B & Instruct & Qwen1.5-7B-Chat & 17.62 & 43.71 & 22.38 & 6.27 & 7.05 & 4.64 & 21.68 \\
7B & Base & Qwen2-7B & 23.93 & 31.49 & 34.71 & 20.39 & 7.27 & 14.32 & 35.37 \\
7B & Instruct & Qwen2-7B-Instruct & 27.94 & 56.79 & 37.81 & 27.64 & 6.38 & 7.37 & 31.64 \\
7B & Instruct & Mistral-7B-Instruct-v0.1 & 12.77 & 44.87 & 7.65 & 2.27 & 0.00 & 6.13 & 15.72 \\
7B & Instruct & Mistral-7B-Instruct-v0.2 & 18.51 & 54.96 & 22.91 & 3.02 & 3.47 & 7.61 & 19.08 \\
7B & Instruct & Mistral-7B-Instruct-v0.3 & 19.23 & 54.65 & 25.57 & 3.85 & 3.91 & 4.30 & 23.06 \\
7B & Base & Mistral-7B-v0.1 & 14.58 & 23.86 & 22.02 & 2.95 & 5.59 & 10.68 & 22.36 \\
7B & Base & Mistral-7B-v0.3 & 14.23 & 22.66 & 24.04 & 3.02 & 5.59 & 8.36 & 21.70 \\
8B & Base & Meta-Llama-3-8B & 13.63 & 14.55 & 24.50 & 4.53 & 7.38 & 6.24 & 24.55 \\
8B & Instruct & Meta-Llama-3-8B-Instruct & 23.91 & 74.08 & 28.24 & 8.69 & 1.23 & 1.60 & 29.60 \\
8B & Base & Llama-3.1-8B & 14.42 & 12.46 & 25.30 & 6.57 & 8.05 & 8.72 & 25.42 \\
8B & Instruct & Llama-3.1-8B-Instruct & 23.76 & 49.22 & 29.38 & 15.56 & 8.72 & 8.61 & 31.09 \\
8B & Instruct & Ministral-8B-Instruct-2410 & 24.19 & 58.96 & 25.82 & 19.56 & 4.59 & 10.72 & 25.46 \\
8B & Base & Qwen3-8B\tnote{$\dagger$} & -- & -- & 78.40 & 60.80 & 44.44 & -- & 56.73 \\
9B & Base & gemma-2-9b & 21.21 & 20.40 & 34.10 & 13.44 & 10.51 & 14.30 & 34.48 \\
9B & Instruct & gemma-2-9b-it & 32.07 & 74.36 & 42.14 & 19.49 & 14.77 & 9.74 & 31.95 \\
\bottomrule
\end{tabular}
\end{adjustbox}
\begin{tablenotes}
\small
\item[$\dagger$] Qwen3 base models evaluated using metrics reported in the Qwen3 technical report~\cite{qwen32025}; IFEval, MUSR, and Average were not reported and are marked `--'.
\end{tablenotes}
\end{threeparttable}
\end{table}

\subsubsection{\textbf{Mistral:}}
Mistral 7B demonstrates superior performance over the previous 13B model (Llama 2) in multiple tasks, and the 34B model (LLaMa 34B) in mathematics and code generation. Incorporating grouped-query attention (GQA) and sliding window attention (SWA) with a rolling buffer cache, Mistral 7B outperforms Llama in reasoning, comprehension and other tasks as evident in benchmarks on datasets such as MMLU, HellaSwag, WinoG, PIQA, Arc-e, Arc-c, NQ, TriviaQA, HumanEval, MBPP, MATH, and GSM8K \cite{Zellers2019HellaSwagCA,Mihaylov2018CanAS, Bisk2019PIQARA,Hendrycks2020MeasuringMM,Zheng2023CodeGeeXAP}. Depending on the task, Mistral's effective size can reach up to 38B. Derivatives of the Mistral 7B have also shown great promise. Zephyr 7B used Direct Preference Optimization (DPO) with the Ultrachat and Ultra-feedback datasets to create a model that outperforms Mistral in MT-bench and Alpaca eval \cite{tunstall2023zephyr,Rafailov2023DirectPO,Ding2023EnhancingCL,Cui2023UltraFeedbackBL}. A sparse mixture of experts derivative of Mistral ($8\times7$B parameters) was shown to outperform Llama2-70B and GPT-3.5 at several tasks \footnote{https://arxiv.org/abs/2401.04088}. At the time of this writing, 8 out of the 10 top performing models on the Open LLM leaderboard on Huggingface were Mixtral derivates (the other two were Llama derivatives). At the time of this writing, Eagle 7B, a model trained on the RWKV architecture outperformed all 7B models including Mistral 7B on cross-lingual benchmarks \cite{peng2023rwkv}.

Latest variants of Mistral's small language models, Ministral 3B and 8B incorporate large proportions of multilingual and code data while featuring a interleaved sliding-window attention pattern that differs from traditional sequential attention mechanisms. This sliding-window approach enables linear compute cost scaling and exploits stacked transformer layers to access information beyond the immediate window size through hierarchical attention patterns. Performance benchmarks demonstrate Ministral 3B achieving 60.9 on MMLU compared to Gemma 2 2B (52.4) and Llama 3.2 3B (56.2), while Ministral 8B scored 65.0 versus Llama 3.1 8B's 64.7. Notably, Ministral 3B outperforms the larger Mistral 7B on most benchmarks despite having fewer parameters, demonstrating significant efficiency improvements in knowledge, reasoning, and function-calling capabilities.

\subsubsection{\textbf{Phi:}}
\label{sec:phi}
The Phi series of models developed by Microsoft started with the Phi-1 focusing on code generation \cite{gunasekar2023textbooks}. The dataset used to train the Phi-1 models, totaling about 7B tokens, is composed of: a filtered code-language dataset, primarily from The Stack and StackOverflow, refined using a language model-based classifier (approximately 6B tokens); a synthetic textbook dataset comprising under 1 billion tokens of Python textbooks generated by GPT-3.5 (as per GPT4 technical report); and a smaller set of synthetic exercises, including around 180 million tokens of Python exercises and solutions. The authors showed that the quality of training data is of significant value compared to the total quantity in billions of tokens. Specifically Phi-1, a 1.3B parameter model trained with 7B tokens, outperformed much larger models such as the Codex-12B (trained with 100B tokens), CodeGen-Mono-16.1B (577B tokens),  PaLM-Coder-540B (780B tokens), and GPT3.5 175B (dataset size unknown) \cite{Chen2021EvaluatingLL,Nijkamp2022CodeGenAO,chowdhery2023palm}. 

The authors extended their work to include common sense reasoning and language understanding with Phi-1.5, an approximately 1.3B parameter model with a dataset that extended the previous Phi-1's data with another 20B tokens of synthetically generated "textbook quality" data. The phi-1.5 model achieved results comparable to larger Llama2 (7B), Falcon (7B) and Vicuna (13B) on benchmarks tested such as WinoGrande, ARC-Easy, ARC-Challenge, BoolQ and SIQA. Continuing this work, the authors augmented their training corpus with select web data, prioritizing educational value and content quality. They scaled up from the 1.3 billion parameter model, Phi-1.5, to the 2.7 billion parameter Phi-2, embedding Phi-1.5's knowledge into Phi-2. This scaling strategy accelerated training and enhanced Phi-2's benchmark performance to a level close to the Llama-2 70B when tested across common sense reasoning, language understanding, math and coding tasks. Phi-3 is a family of small models from 2.8B to 14B parameters. The Phi3-mini, a 3.8 billion parameter language model, competes with larger models like Mixtral 8x7B and GPT-3.5, achieving 69\% on MMLU and 8.38 on MT-bench (as per phi-3 technical report). Phi-3's training dataset is an enhanced version of the one used for phi-2, comprising filtered web data and synthetic data, ensuring robustness, safety, and chat format alignment. Phi-4 (as per phi-4 technical report) outperforms its predecessor phi-3 (14B) across all evaluated benchmarks, achieving notable improvements in MMLU (84.8 vs. 77.9), GPQA (56.1 vs. 31.2), and MATH (80.4 vs. 44.6). When compared to Qwen 2.5 (14B instruct), phi-4 delivers higher scores on critical reasoning tasks such as MMLU, GPQA, and MATH, showcasing its superior ability to handle complex reasoning challenges. Additionally, phi-4 surpasses GPT-40-mini in mathematical reasoning (MATH) and general benchmarks like MGSM, further solidifying its competitive edge within the small model category.

\subsubsection{\textbf{Orca:}}
In the Orca model series, authors found that using synthetic data from LLMs, like the Phi models, improved benchmark performance but didn't enhance reasoning skills, as the models primarily learned the style and answers of LLMs (microsoft research technical report). In contrast to the Phi line of models, Orca models were trained using explanation tuning where system instructions complement user inputs, guiding the system to generate well-reasoned responses. These instructions used methods like chain-of-thought and simplification strategies to use this data to enable smaller models to emulate GPT-4's thinking process using pairs of system and user instructions with inputs and outputs. Utilizing the FLAN-v2 data, the study sampled 5 million user queries for ChatGPT responses and further selected 1 million from these for GPT-4 responses to ensure a large and diverse dataset \cite{Longpre2023TheFC}. For both tasks testing reasoning abilities (AGIeval and BigBench), Orca (13B) retained up to $88\%$ performance of ChatGPT, outperforming other models like the Vicuna 13B \cite{Srivastava2022BeyondTI,Zhong2023AGIEvalAH}. Moreover, Orca reaches parity with ChatGPT on the BBH benchmark and shows competitive performance (4 pts gap with optimized system message) in professional and academic examinations like the SAT, LSAT, GRE, and GMAT, both in zero-shot settings without CoT; while trailing behind GPT-4.

In the authors' derivative work with Orca 2, the model is trained with 
various reasoning techniques (step-by-step, recall then generate, recall-reason-generate, direct
answer, etc.) with the aim of determining the best solution strategy for each task. The Orca 2 dataset extends Orca 1 dataset with ~817K new training instances. The training starts with the Llama2-7B or 13B model, and progressively trains with the Flan, Orca 1 and Orca 2 datasets. Benchmarks for testing reasoning abilities, math, knowledge understanding, safety and truthfulness showed that the Orca 2 model outperformed models several times its size including the WizardLM (70B) and Llama2 chat (70B) \cite{Mitra2023Orca2T}.

\subsubsection{\textbf{Gemini:}}

The Gemini series of multimodal models developed at Google are trained jointly on image, audio, video, and text data to create a lineup of models with strong generalist abilities across different modalities. Gemini models can process textual input combined with a variety of audio and visual inputs, including natural images, charts, screenshots, PDFs, and videos, capable of generating both text and image outputs. The multimodal and multilingual dataset used for training includes data from web documents, books, code, as well as image, audio, and video data resulting in the Gemini family of four models with varying sizes: Gemini Ultra, Gemini Pro, Nano-1, and Nano-2, with a particular emphasis on the smaller models, Nano-1 and Nano-2. Our focus here are the smaller Nano-1 and Nano-2 variants. Nano-1 and Nano-2 are relatively compact, with 1.8 billion and 3.25 billion parameters, respectively. Despite their smaller size, these models exhibit exceptional performance in tasks related to factuality and retrieval, as well as considerable capabilities in reasoning, STEM, coding, and multimodal and multilingual tasks such as BoolQ, Natural Questions, Big Bench, and MMLU \footnote{https://arxiv.org/abs/2312.11805}. The Nano models achieve their proficiency by distilling knowledge from the larger Gemini models. They are 4-bit quantized for efficient deployment, offering the ability to be deployed on mobile and edge devices. Beyond text understanding tasks that have been covered by other models in this section, the Gemini Nano-1 model also surpasses both USM and Whisper in various datasets, with the exception of FLEURS \cite{Radford2022RobustSR,Conneau2022FLEURSFL}.

Gemma 2B and Gemma 7B are open SLMs from Google built from the research and technology used to create Gemini models. Gemma models have been trained on primarily English data from web documents, mathematics, and code. The 2B model was trained on 3 trillion tokens, while the 7B model was trained on 6T tokens. Gemma, which outperforms similarly sized open models on 11 out of 18 common text-based tasks also mentione above for Gemini models.\footnote{https://arxiv.org/abs/2403.08295}

Gemma 2 introduced significant architectural enhancements with a hybrid attention mechanism that interleaves local sliding window attention (4096 tokens) on odd layers with global attention (8192 tokens) on even layers. The models utilize Grouped-Query Attention \cite{ainslie2023gqa} with groups = 2, showing increased inference speed while maintaining downstream performance, combined with post-norm and pre-norm RMSNorm for training stability. Logit soft-capping prevents excessive logit growth by scaling to fixed ranges (50.0 for attention layers, 30.0 for final layers). Knowledge distillation was employed for the 2B and 9B variants, training smaller models to mimic larger teacher models rather than using standard next-token prediction. Performance benchmarks (MMLU, GSM8K, ARC-C, etc.) demonstrate Gemma 2-9B outperforming Llama 3 8B while maintaining computational efficiency suitable for single GPU deployment.

\subsubsection{\textbf{SmolLM2}}
SmolLM2 represents Hugging Face's latest contribution to efficient small language models, available in three sizes: 135M, 360M, and 1.7B parameters\cite{benallal2025smollm2}. SmolLM2 models are trained on 11 trillion tokens using a multi-stage process that strategically combines web text with specialized data. This approaches utilizes AdamW optimizer with a Warmup Stable Decay (WSD) learning rate schedule for training efficiency and uses tokenizer with a vocabulary size of 49,152 tokens. The key innovation in SmolLM2's development is the multi-stage training methodology where dataset mixing rates are manually refined between stages based on performance observations, with the final stage focusing on high-quality mathematical and code data. The researchers created three new specialized datasets to address limitations in existing public resources: FineMath (up to 54B tokens of high-quality mathematical content), Stack-Edu (educational filtering of code from StarCoder2Data), and SmolTalk (instruction-tuning dataset combining existing and new synthetic data). 
Evaluation results show that SmolLM2 base model, outperforms Qwen2.5 base model on multiple benchmarks (example: HellaSwag, ARC, etc.). SmolLM2 also delivers strong performance on held-out benchmarks not monitored during training, such as MMLU-Pro and Natural Questions. Notably, the model outperforms Qwen2.5-1.5B by nearly 6 percentage points on MMLU-Pro, further validating its generalization capabilities. On math and coding benchmarks, SmolLM2 base model demonstrates competitive performance. While it lags behind Qwen2.5-1.5B, SmolLM2 outperforms Llama3.2-1B on GSM8K, MATH and HumanEval. The post-trained SmolLM2-Instruct model demonstrates strong performance across multiple benchmarks following supervised fine-tuning on the SmolTalk dataset and alignment via DPO \cite{Rafailov2023DirectPO} with UltraFeedback \cite{Cui2023UltraFeedbackBL}. The model achieves 56.7\% on IFEval, outperforming both Llama3.2-1B (53.5\%) and Qwen2.5-1.5B (47.4\%) in instruction-following tasks. On mathematical reasoning benchmarks, SmolLM2-Instruct shows competitive results with 48.8\% on GSM8K (5-shot) and 21.0\% on MATH (4-shot), surpassing Llama3.2-1B but falling behind Qwen2.5-1.5B on GSM8K (63.3\%). These results position SmolLM2-Instruct as a balanced performer across diverse tasks, exhibiting particular strengths in instruction-following while showing areas for improvement in certain mathematical and specialized coding tasks

\subsubsection{\textbf{IBM Granite}}
IBM Granite 3.2 dense models (2B, 8B) trained on up to 12 trillion tokens and MoE variants (ex: 3B-A800M) trained on 10 trillion tokens utilize Grouped Query Attention, RoPE, SwiGLU activation, and RMSNorm architectures (as per IBM Granite 3 technical report). The models achieve competitive performance against larger counterparts through inference scaling techniques applied to chain-of-thought reasoning, enabling toggleable reasoning modes that can be activated task-specifically. IBM reports that Granite 3.2-8B demonstrates performance rivaling GPT-4o-0513 and Claude 3.5-Sonnet on MATH500 and AIME2024 benchmarks while maintaining significantly lower computational requirements. The reasoning capabilities were developed through a combination of inference scaling and supervised fine-tuning on diverse instruction-following datasets, allowing the models to generate intermediate reasoning steps before producing final answers

\subsubsection{\textbf{DeepSeek R1}}
DeepSeek R1 \footnote{https://arxiv.org/abs/2501.12948} applies large-scale reinforcement learning directly to base models without supervised fine-tuning as a preliminary step. The original DeepSeek R1 utilizes a mixture-of-experts architecture with 671 billion total parameters but only 37 billion activated per token, built upon the DeepSeek-V3-Base foundation. To make these reasoning capabilities accessible, DeepSeek released a comprehensive suite of distilled models ranging from 1.5B to 70B parameters, created by fine-tuning existing architectures including Qwen 2.5 and Llama 3 series on reasoning data generated by the full DeepSeek R1 model. The distillation process involved supervised fine-tuning on 800k samples of reasoning chains generated by DeepSeek R1, enabling smaller models to capture complex reasoning patterns. Performance results demonstrate that DeepSeek-R1-Distill-Qwen-1.5B outperforms GPT-4o and Claude-3.5-Sonnet on math benchmarks with 28.9\% on AIME and 83.9\% on MATH. These distilled models show that sophisticated reasoning capabilities developed through reinforcement learning in large models can be effectively transferred to smaller, more practical architectures suitable for deployment on consumer hardware while preserving much of the original reasoning performance.

\begin{figure}[h!]
\includegraphics[width=\linewidth, trim=0 0 0 3cm, clip]{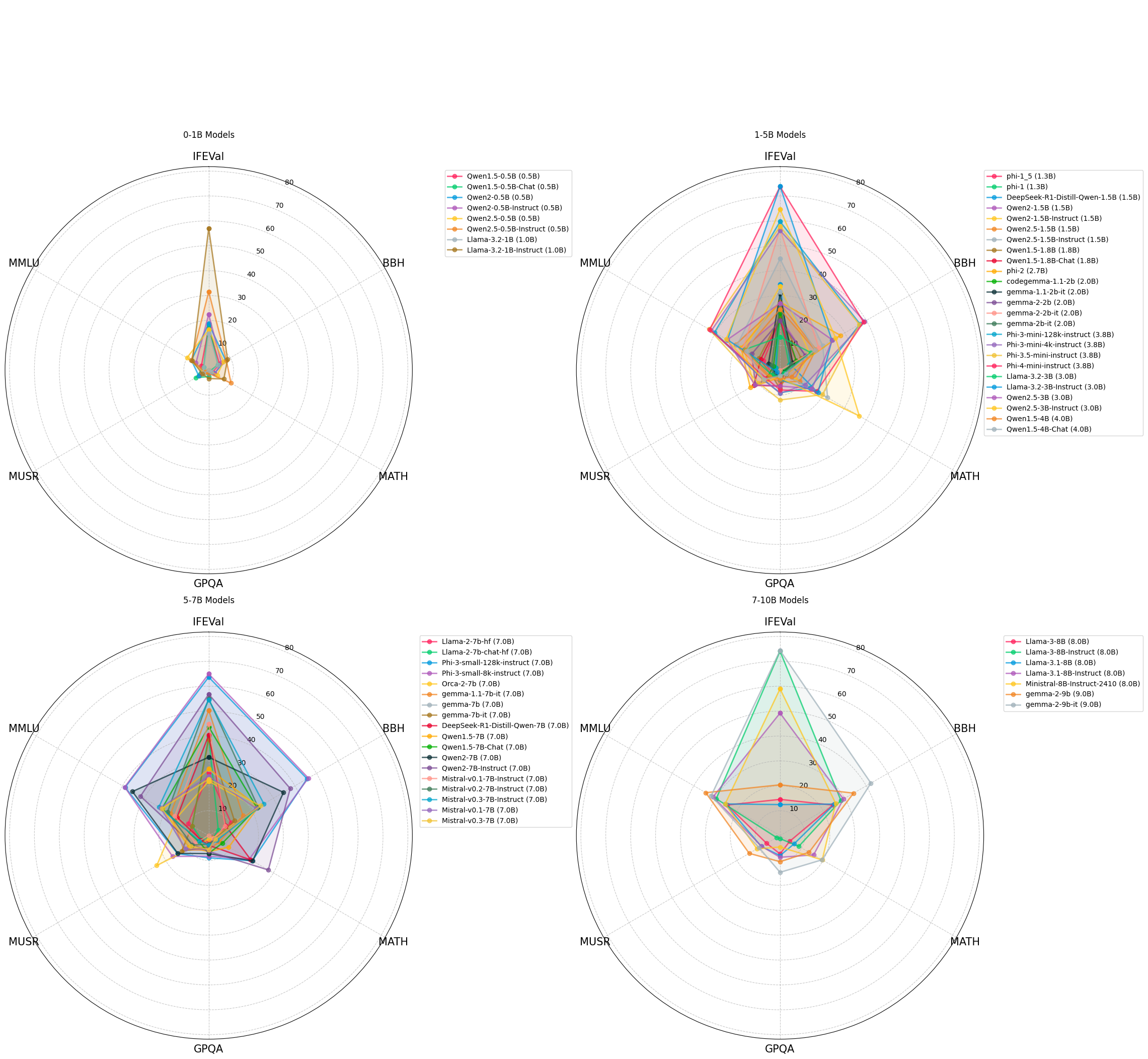}
\vspace{-0.7cm}
\caption{Performance comparison by model size and benchmark dataset}
\vspace{-0.1cm}
\label{fig:radar}
\end{figure}

\subsubsection{\textbf{Hybrid State Space Models and Efficient Architectures:}} Recent innovations in neural architectures have focused on addressing three critical challenges: reducing memory footprint, improving computational efficiency, and maintaining or enhancing model performance. State space models have emerged as a promising alternative to attention mechanisms, offering linear computational complexity and efficient parameter usage\cite{gu2023mamba}. However, the integration of SSMs into practical architectures presents its own set of challenges, particularly in maintaining the strong performance characteristics of attention-based models while achieving better efficiency.

In this section, we examine recent architectural innovations that represent significant departures from pure transformer models. We focus on four representative architectures that showcase different approaches to combining SSMs with other mechanisms: parallel fusion (Hymba)\cite{dong2024hymba}, sequential fusion with shared parameters (Zamba)\footnote{https://arxiv.org/abs/2405.16712}, interleaved fusion with mixture-of-experts (Jamba)\cite{lieber2024jambahybridtransformermambalanguage}, and pure selective SSMs (Mamba)\cite{gu2023mamba}.

Hymba\cite{dong2024hymba} introduces a hybrid architecture combining transformer attention with state space models (SSMs) in parallel within each layer. It uses meta-tokens and cross-layer KV cache sharing to improve efficiency.
Zamba\cite{glorioso2024zambacompact7bssm} uses a Mamba backbone (SSM-based) with a single shared global attention module repeated every few layers. The shared attention reduces parameters while maintaining performance.
Jamba\cite{lieber2024jambahybridtransformermambalanguage} interleaves transformer and Mamba layers with mixture-of-experts (MoE) to increase model capacity while keeping active parameters manageable.
Mamba\cite{gu2023mamba} introduces selective SSMs that allow input-dependent parameter updates, combined with hardware-optimized parallel scan algorithms. It uses a simplified architecture without attention or MLPs.

The key innovation of each approach is unique where Hymba focuses on efficient parallel fusion, Zamba on shared attention, Jamba on MoE integration, and Mamba on selective SSMs with hardware optimization. Each makes different trade-offs between model capacity, efficiency and architectural complexity and the comparisons among these models are provided in Table \ref{table:model_comparison}.

\subsection{Task Specific SLMs}
\label{sec:tsslm}
Recent techniques seek to teach SLMs to employ different solution strategies for different tasks, potentially different from the one used by the larger model. For example, Orca \footnote{https://arxiv.org/abs/2311.11045} learns various reasoning techniques (step-by-step, recall then generate, recall-reason-generate, direct answer, etc.) that determine the most effective solution strategy for each task. Various prompting techniques like step-by-step Chain-of-Thought (CoT) has proved to be effective in task specific LMs, when combined with knowledge distillation \cite{shridhar2023distilling, schick2021its, Magister2022TeachingSL}. We discuss some of the task specific SLMs in this section. 

\subsubsection{\textbf{Mathematical Reasoning:}} 

Specializing SLMs for math reasoning tasks with multi-step CoT has proven effective \cite{fu2023specializing}. Techniques such as Equation-of-Thought Distillation (EoTD), Mix Thoughts Distillation (MTD) \footnote{https://arxiv.org/abs/2401.11864}, and CogTree \cite{yan2023complex} enhance reasoning capabilities. WizardMath 7B surpasses most 7B-40B open-source models, while WizardMath 13B exceeds Llama 2 70B on GSM8k \cite{luo2023wizardmath}.

\subsubsection{\textbf{Code Generation:}}

WizardCoder \cite{luo2023wizardcoder},  a Code Evol-Instruct fine-tuned Code model, achieves superior performance compared to Anthropic's Claude and Google's Bard and surpasses other open-source code LLMs on four benchmarks HumanEval, HumanEval+, MBPP, and DS-1000. Code Llama \footnote{https://arxiv.org/abs/2308.12950}  is a family of Llama 2 where Code Llama Python 7B outperforms Llama 2 70B
on HumanEval and MBPP. Stable Code 3B is a 3B parameter code completion model that achieves on par performance on Multi-PL as compared to Code Llama 7b, as per Stability AI technical report. 

\subsubsection{\textbf{Code Decompilation:}}

SLaDe \cite{armengolestapé2024slade}, 200M parameter transformer based model trained to decompile assembly level code to C code using novel code tokenization and dropout free regularization. Large-scale evaluations on over 4000 executable programs from ExeBench, AnghaBench, neural network decompiler BTC and state of art industrial strength decompiler Ghidra and LLM ChatGPT show SLaDe significantly outperform despite many order magnitude fewer weights producing 1.17x to 3.83x more correct code, and a higher edit similarity than Ghidra, ChatGPT and BTC.

\subsubsection{\textbf{Machine Translation:}}

ALMA-13B-R \cite{xu2024contrastive} is a 13B translation model which stands out as the first moderate-size LLM-based translation model that surpasses GPT-4 and WMT competition winners in neural machine translation task. ALMA family of SLMs are based on LLaMA-2 and the performance is significantly better than all prior work and even superior to the NLLB-54B model and GPT3.5-text-davinci-003, with only 7B or 13B parameters \cite{xu2024paradigm}. ALMA-13B-R is a new variation that originated from ALMA where it uses  Contrastive Preference Optimization that trains models to avoid generating adequate but not perfect translations. This has further improved the performance against GPT4 on WMT’21, WMT’22 and WMT’23 datasets. In another study, small-sized pre-trained language model outperformed the extra-large language models in clinical domain fine-tuning \cite{han2023neural}. The models achieved top-level performances in the ClinSpEn-2022 shared task on English-Spanish clinical domain data.

\subsubsection{\textbf{Function Calling:}}

Function calling represents a rapidly emerging SLM specialization for interacting with external APIs and executing programmatic functions, valuable for automation and software development.

The Octopus series pioneered on-device function calling without cloud-based processing. Octopus v1 \cite{chen2025octopus} introduced task-oriented models specifically designed for API function calling, achieving 99.524\% accuracy and outperforming GPT-4 (98.571\%). The model employs conditional masking techniques to ensure outputs conform to required API formats, addressing a critical challenge in function calling accuracy. Octopus v2\footnote {https://arxiv.org/abs/2404.01744} advanced this by incorporating functional tokens directly into the tokenizer, combining function selection and parameter generation into a unified process. This architectural innovation achieved inference times of 0.36-0.38s compared to GPT-4 (1.02s) and Llama-7B-RAG (13.46s), demonstrating significant efficiency improvements.

TinyAgent \cite{erdogan2024tinyagent} presents an end-to-end framework employing the LLMCompiler framework with systematically curated datasets to fine-tune models ranging from 1.1B to 7B parameters. TinyAgent achieved function calling success rates of 80.06\% (1.1B) and 84.95\% (7B), outperforming GPT-4-Turbo's 79.08\%. The framework enables edge deployment with enhanced privacy, reduced latency, and lower operational costs, demonstrating that SLMs can achieve competitive performance while maintaining deployment advantages.

The Granite Function Calling Model \cite{abdelaziz2024granite} introduces a multi-task learning approach that decomposes function calling into granular sub-tasks including function identification, argument parsing, and return value handling. This decomposition enables SLMs to develop robust understanding of the complete function calling process. Granite-20B achieved 8\% better F1 score than competing models on function name detection, with 7\% improvement on LCS and 11\% on Exact Match scores \cite{abdelaziz2024granite}.

ToolACE \cite{liu2025toolace} enhances capabilities through advanced data generation pipelines. Function calling SLMs face challenges including performance variations across benchmarks and handling multi-step reasoning chains. Recent work \footnote{https://arxiv.org/abs/2501.10132} explores multi-step and constrained scenarios, with RAG integration showing promise for enhancing accuracy. Future directions include robust multi-step handling, improved reliability across diverse API types, and multilingual capabilities.

\subsubsection{\textbf{Multi-modal Task:}}
Multi-modal small language models (SLMs) have emerged as specialized architectures that efficiently process and generate content across different modalities while maintaining computational efficiency.

MiniGPT-4 represents a breakthrough in efficient vision-language understanding, developed by researchers at King Abdullah University of Science and Technology \cite{minigpt4}. The model demonstrates that smaller architectures can achieve strong vision-language capabilities by leveraging pre-trained vision encoders combined with large language models through a single projection layer. The approach focuses on aligning visual features with language representations using approximately 5 million aligned image-text pairs, requiring only the training of a lightweight projection layer while keeping both the vision encoder and language model frozen. MiniGPT-4 outperformed BLIP-2 particularly in creative tasks, successfully handling 80\% of recipe, advertisement, and poem generation tasks.

LLaVA (Large Language and Vision Assistant) and its improved version LLaVA-1.5 have established new benchmarks for efficient multi-modal reasoning \cite{llava}. Developed by researchers from the University of Wisconsin-Madison and Microsoft, LLaVA demonstrates that models in the 7B-13B parameter range can achieve competitive performance on vision-language benchmarks through visual instruction tuning. The architecture efficiently combines a vision encoder (CLIP ViT-L/14) with language models (Vicuna) through learnable projection layers, enabling effective multi-modal understanding with careful architectural design. LLaVA-1.5-13B achieves 90.92\% accuracy on ScienceQA, 85.1\% relative score compared to text-only GPT-4 on LLaVA-Bench, and 81.7\% on complex reasoning tasks.

InstructBLIP advances instruction-following vision-language models by extending the BLIP-2 architecture with instruction-aware components \cite{instructblip}. Developed by researchers from Salesforce, Hong Kong University of Science and Technology, and Nanyang Technological University, InstructBLIP introduces an instruction-aware Query Transformer (Q-Former) that takes instruction text tokens as additional input, enabling the extraction of task-relevant image features and allowing 7B-13B models to compete effectively with larger proprietary models.

MoE-LLaVA\footnote{https://arxiv.org/abs/2401.15947} employs mixture-of-experts architecture to handle diverse multi-modal tasks efficiently. MoE-LLaVA-1.8B×4 with only 2.2B activated parameters surpasses LLaVA-1.5-13B by 1.0-1.5\% on object hallucination benchmarks, demonstrating that sparse activation patterns can enable smaller models to achieve competitive performance.

VideoChat \cite{videochat} extends multi-modal capabilities to video understanding, showing that 7B models can achieve competitive performance on video question answering tasks through integrating video foundation models with large language models via learnable neural interfaces.
\begin{table}[h]
\centering
\caption{Performance comparison of multi-modal small language models on key benchmarks}
\label{tab:mslm_performance}
\begin{tabular}{|l|c|c|c|c|c|}
\hline
\textbf{Model} & \textbf{Parameters} & \textbf{ScienceQA} & \textbf{LLaVA-Bench} & \textbf{POPE} & \textbf{Complex Reasoning} \\
\hline
MiniGPT-4 & 7B & - & 2.22 (GT captions)$^*$ & - & 80\% (creative tasks) \\
LLaVA-1.5-7B & 7B & - & - & - & - \\
LLaVA-1.5-13B & 13B & 90.92\% & 85.1\%$^{**}$ & 86.6\% & 81.7\% \\
InstructBLIP-13B & 13B & - & - & - & Superior ClipM/EM \\
MoE-LLaVA-1.8B×4 & 2.2B (activated) & - & - & 87.4\% & - \\
\hline
\end{tabular}
\begin{tablenotes}
\small
\item $^*$ Average ground truth captions compared to BLIP-2's 1.96
\item $^{**}$ Relative score compared to text-only GPT-4
\end{tablenotes}
\end{table}

The success of these multi-modal SLMs demonstrates that domain-specific architectural innovations, efficient training strategies, and careful model design can significantly enhance performance-to-parameter ratios. However, challenges remain in handling complex reasoning across modalities and maintaining consistent performance across diverse multi-modal tasks.

\subsubsection{\textbf{Other Task Specific SLMs:}}
FLAME \cite{joshi2023flame}, a 60M parameter T5-based model for Excel formulas, outperforms larger models (Codex-Davinci 175B, Codex-Cushman 12B) in 6 of 10 settings using two orders of magnitude less training data. Prometheus \cite{kim2023prometheus,lee2024prometheus} demonstrates growing interest in SLMs as alignment judges. Competitions like BabyLM \cite{conll-2023-babylm} are driving further task-specific SLM development.

While the focus of this paper is SLMs, we also note that similar advancements are seen in other modalities like text-to-image generation. For example, Segmind Stable Diffusion 1B (SSD-1B) is a distilled version of the Stable Diffusion XL(SD XL) which offers $60\%$ speedup while closely mimicking the output of the base model(SDXL). Techniques used in Segmind, such as architectural compression and feature distillation are also relevant to SLMs. Other useful techniques transferable to SLMs include speeding up image generation through the use of modules like SpeedupNet and LCM-LoRA which fuses the acceleration capabilities of Latent Consistency Models with LoRA \cite{chai2023speedupnet}.

\subsection{Domain Specific SLMs}

While task agnostic SLMs offers a broad range of knowledge and reasoning capabilities, industry vertical-based SLMs excel in higher accuracy in specific contexts and efficient for industry-specific tasks. We discuss some of the vertical specific SLMs in this section.

\subsubsection{\textbf{Medical Domain:}} 

{BioGPT}\footnote{https://arxiv.org/abs/2305.07804} is a medical SLM fine-tuned on PubMedQA dataset\cite{jin2019pubmedqa}, using generative data augmentation, outperforming few-shot GPT-4. It effectively used Low-Rank Adaptation (LoRA) for domain-specific adaptation. Other medical SLM papers also focus on data augmentation using LLMs.\cite{zhou2021datlmedqa,pappas2022data}.

Meerkat-7B is an open 7B-parameter medical SLM (built on Mistral-7B) for clinical knowledge and multi-step reasoning. It was fine-tuned on a ~460k-example dataset, including high-quality chain-of-thought solutions to medical questions. Specifically, this involved extracting reasoning paths for USMLE-style questions and synthesizing Q\&A pairs from 18 medical textbooks \cite{lee2025meerkat}. This instruction tuning imparted strong problem-solving skills. Meerkat-7B achieved state-of-the-art results among 7B models, averaging 64.5\% accuracy on expert test sets and surpassing MediTron-7B and BioMistral-7B. Notably, it scored 77.1 on MedQA USMLE, exceeding the 60\% passing threshold and outperforming GPT-3.5 \cite{lee2025meerkat}. These results show a small, domain-specific model with enriched reasoning can rival larger models.

RiskAgent is an 8B-parameter autonomous medical AI copilot developed by researchers at the University of Oxford and University College London. Based on LLaMA-3-8B, it operates as a multi-agent system (Decider, Executor, Reviewer) that is explicitly trained to collaborate with an external library of 387 evidence-based clinical decision tools, such as risk calculators and scoring systems. Rather than internalizing all medical knowledge, RiskAgent learns to select and execute the appropriate tool for a given problem. On the novel MedRisk benchmark, RiskAgent-8B achieves 76.33\% accuracy, more than doubling the performance of GPT-4o and showcasing particular strength in rare disease scenarios. This approach produces highly accurate and trustworthy results with traceable information sources, making it a robust and resource-efficient solution for clinical applications.

ClinicalGPT-R1 \footnote {https://arxiv.org/abs/2504.09421v2} is a reasoning-enhanced language model designed for bilingual (Chinese and English) disease diagnosis. It was trained on a dataset of 20,000 real-world clinical records, for which the diagnostic reasoning paths were synthesized by state-of-the-art LLMs. The model was then developed using a two-stage process of supervised fine-tuning (SFT) and reinforcement learning (RL) to enhance its clinical reasoning capabilities. Its performance was evaluated on MedBench-Hard, a challenging, custom-curated dataset, highlighting a key trend of leveraging powerful LLMs as data synthesizers to train smaller, more specialized models

\subsubsection{\textbf{Finance Domain:}}

FinGPT \cite{yang2023fingpt}, an open-source language model for the finance, provides researchers and practitioners with accessible and transparent resources to develop FinLLMs. FinGPTs potential applications include robo-advising, algorithmic trading, and low-code development, which can be seen as stepping stones for users. SLMs have been explored for task-specific training (like FinBERT) but pretraining and instruction fine-tuning have only been explored for large models over the 65B range (InvestLM, BloombergGPT\footnote{https://arxiv.org/abs/2303.17564}  \cite{liu2021finbert}. The decision process to select LLMs over SLMs for this domain was briefly explained in \cite{li2023large}.

FinBloom 7B \footnote{https://arxiv.org/abs/2502.18471} a BLOOM-based financial SLM, combines domain-adaptive pretraining (26M financial documents) with instruction tuning (50k FinContext dataset). It achieves state-of-the-art performance on FinBench, a suite of 35 financial NLP tasks. FinBloom’s grounding mechanism retrieves real-time financial news or SEC filings, allowing it to outperform models like FinGPT in dynamic market scenarios. This distinguishes FinBloom for applications requiring up-to-date interpretation. However, its reliance on static grounding and English-only training are limitations. Future directions include live tool augmentation and multilingual tuning. The open-source model has use cases from regulatory Q\&A to market intelligence.

\subsubsection{\textbf{Legal Domain:}} 

Lawyer LLaMA is one of the earliest attempts of building LLMs in Chinese legal domain. Authors propose injecting domain knowledge during the continual training stage and designing proper supervised fine-tuning tasks to help the model tackle practical issues (as per Lawyer LLaMa technical report). ChatLaw is another Chinese legal domain expert model which is LoRA fine-tuned version of Ziya-LLaMA-13B \cite{zheng2023judging} on 937k Chinese National Law examples that outperforms both GPT-4 and Lawyer LLaMA. 

LexLM-7B \footnote{https://arxiv.org/html/2407.21065v1} is a domain-specialized 7B-parameter model trained on 1.5 billion tokens of U.S. legal data, including statutes, case law, and contract templates. Fine-tuned with instruction-based supervision on legal entailment and question answering tasks, LexLM-7B achieves 73.2\% on LegalBench QA, outperforming LegalBERT and CaseLaw-BERT by 6–9 points on tasks such as ContractNLI and EntailmentBank-Legal. Its architecture enables clause-level reasoning and legal lookup with structured text inputs, making it suitable for applications in compliance automation, legal research, and contract review.
Despite these capabilities, LexLM’s adaptation reveals limitations in tracking legal precedents, handling multi-jurisdictional variance, and reasoning over case timelines, which often exceed the model’s context length. Moreover, it lacks grounding in real-world legal workflows and is not integrated with citation graphs or legal retrieval tools. Future research could explore graph-based legal memory, temporal reasoning across evolving legislation, and alignment with legal otologies (e.g., EuroVoc, LexML). With appropriate audit and interpret ability mechanisms, LexLM-7B could serve as the foundation for next-generation legal AI systems that balance efficiency and regulatory accountability.

\subsubsection{\textbf{Telecom Domain:}}

TelcoLM \footnote{https://arxiv.org/abs/2412.15891} developed by Orange France, is a 7B-parameter SLM adapted from LLaMA-2 through instruction tuning on 800 million telecom-specific tokens and 80,000 curated prompts. Without additional domain-adaptive pretraining, the model achieves performance on par with GPT-3.5 across TelcoBench, a benchmark covering diagnostics, SLA configuration, and customer support dialog. TelcoLM demonstrates that carefully designed instruction sets targeting troubleshooting, hardware queries, and service flow issues can allow smaller models to compete with general-purpose LLMs. The training pipeline focuses on single-step instruction tuning (IAPT), highlighting the value of curated domain supervision over large-scale token count. While TelcoLM performs strongly within its adaptation domain, the model shows limited cross-domain generalization when exposed to inputs that diverge even slightly from its fine-tuned domain, highlighting a key limitation of narrow adaptation. This underscores the challenge of maintaining robustness when telecom models encounter edge cases, hybrid services, or multidisciplinary queries. Future work may explore multi-domain alignment, live network log ingestion, and modular adapters for multilingual support. TelcoLM’s architecture is well-suited for on-device diagnostics, tier-1 customer support, and interactive service resolution within fixed telecom settings.

\subsubsection{\textbf{Mining Domain:}}

MiningGPT is a 7B-parameter instruction-following model specialized for the mining industry, created by fine-tuning a base Mistral-7B Instruct model on mining-specific data based on a research done around a domain-specific large language model for the mining industry. To build its training corpus, the developers assembled a dataset called MiningPile (~120 million tokens) by filtering large open datasets (e.g. the C4 web corpus and technical papers on arXiv) using ~600 mining-related keywords, then augmenting with domain-specific documents such as mining engineering theses and reports. They used a 100B+ general LLM (Google’s Gemini, in this case) to generate question–answer pairs from the MiningPile texts, producing a synthetic instruction dataset focused on mining operations, terminology, and technical Q\&A. MiningGPT was then fine-tuned (using QLoRA for efficiency) on these domain-specific QAs, resulting in a model that demonstrated a 14\% higher mining-domain knowledge score than the original 7B baseline model. In practice, MiningGPT shows markedly improved understanding of mining jargon, equipment, and processes, outperforming general models on tasks like interpreting drill reports or troubleshooting mining machinery issues \footnote{https://arxiv.org/abs/2412.01189}. As an open research model, it exemplifies the gains achievable through targeted domain adaptation in heavy industry, enabling more accurate and context-aware responses in the mining engineering domain.

\subsubsection{\textbf{Scientific Reasoning Domain:}}

SciPhi-2 is a 5.6B parameter model based on Microsoft’s Phi-2, is designed for high-precision scientific reasoning across physics, chemistry, mathematics, and computer science. It leverages instruction tuning on curated scientific documents, ArXiv abstracts, and symbolic problem sets to achieve 89.1\% on ARC Challenge and 84.6\% on MMLU-STEM, outperforming models nearly twice its size. SciPhi-2 integrates equation-aware decoding and symbolic grounding techniques, enabling accurate performance on logic puzzles, proof verification, and multi-step math word problems. Its architecture reflects an emerging class of compact, domain-focused LLMs with enhanced reliability in scientific inference tasks.

However, SciPhi-2 remains limited in handling multi-modal scientific inputs such as graphs, tables, and plots. The model also under performs on tasks requiring unit conversion, dimensional analysis, or temporal causality—areas common in experimental science. Further research is needed to augment these models with tool-calling abilities (e.g., Wolfram, calculators), multi-hop document linking, and interactive theorem proving capabilities. As scientific workflows become increasingly data-rich and cross-disciplinary, models like SciPhi-2 may evolve into assistant agents for hypothesis generation, technical summarization, and STEM education support, while still demanding improved interpret ability and reproducibility for peer-reviewed environments.

\begin{figure}[h!]

\includegraphics[width=\linewidth, trim=1cm 1cm 0 1cm, clip]{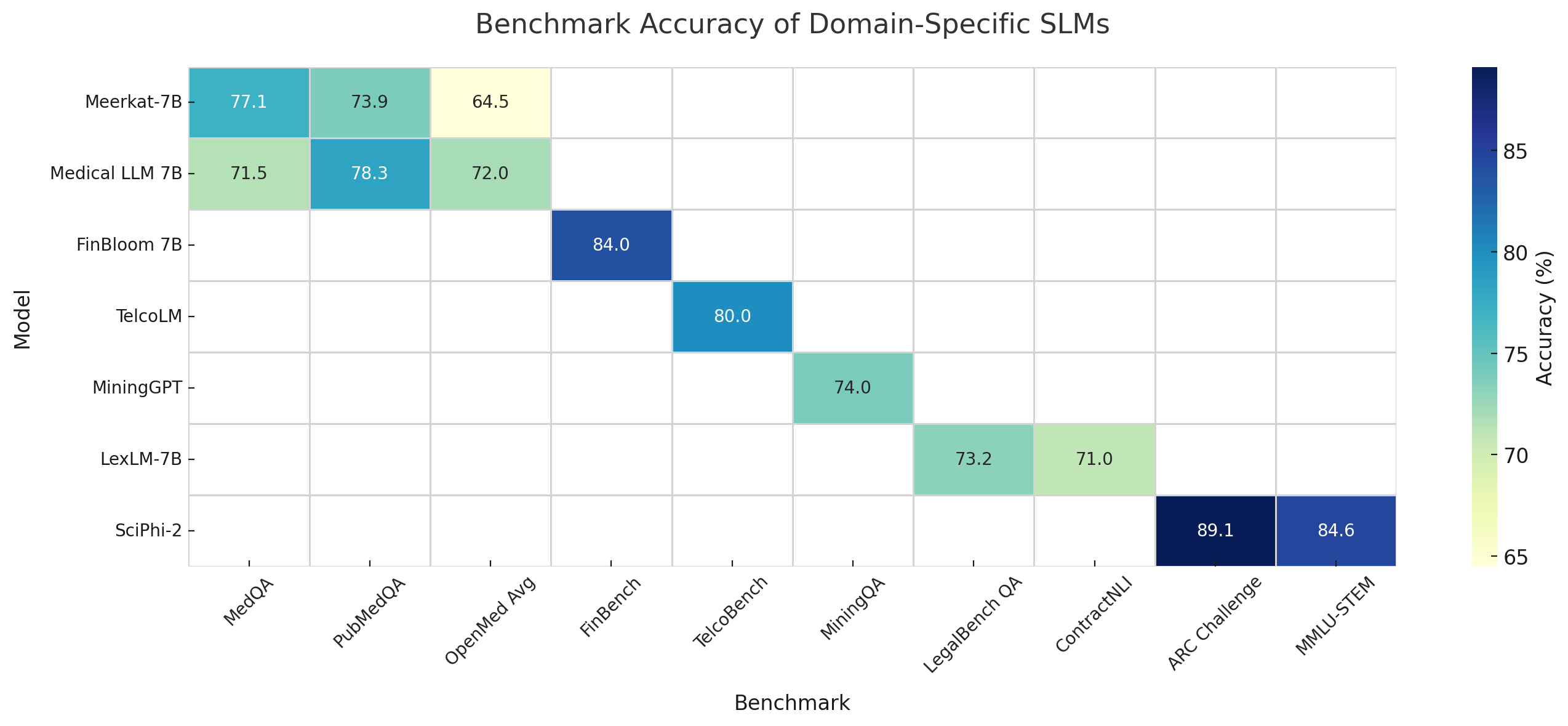}
\vspace{-0.7cm}
\caption{Benchmark Accuracy for domain specific SLMs}
\vspace{-0.1cm}
\label{fig:heatmap}
\end{figure}

\section{Approaches to Create SLMs}

\subsection{Training Techniques}

In this section, we firstly review different innovative methodologies involved in training SLMs including knowledge distillation, instruction tuning, Chain-Of-Thought (CoT) etc, demonstrating efficacy in training smaller parameter models. Then, we focus on recent trend representing a significant shift in how SLMs are structured and optimized for performance and efficiency. We discuss about blending ensemble SLMs and novel way of combining smaller models in mixture of experts (MoE) models later in this section.

\subsubsection{\textbf{Imitation Learning (IL):}}

Initial research in training smaller language model through imitation learning from Large Foundational Models (LFMs) showcased number of issues in the quality of these models. For instance, models like MiniLM \cite{wang2020minilm} and XtremeDistil \cite{mukherjee2020xtremedistil} were trained on homogeneous training data and due to lack of rigorous evaluation techniques resulted in overestimating the small model capabilities. IL fundamentally relies on the quality and diversity of the demonstration data. This reliance presupposes the availability of high-quality, representative examples, which is not always feasible, particularly in complex or novel scenarios where expert demonstrations are scarce or costly to obtain. Secondly, IL suffers from a phenomenon known as the distribution shift. This occurs when the model encounters situations not covered in the training data, leading to a significant degradation in performance due to the model's inability to generalize beyond its training examples. 
    
\subsubsection{\textbf{Progressive Learning:}} 

Orca \footnote{https://arxiv.org/abs/2306.02707}, 13B SLM is trained using a progressive learning approach overcomes limitation of IL, where it learns to imitate the reasoning process of large foundation models (LFMs) such as GPT-4. Technique involves training SLMs using instruction tuning, where a smaller "student" model is trained using output generated by larger foundation model. Orca learns from rich signals from GPT-4, including explanation traces, step-by-step thought processes, and other complex instructions. It is guided by teacher assistance from ChatGPT. Orca2 \footnote{https://arxiv.org/abs/2311.11045}, smaller model effectively mimic the style of teacher model on knowledge-intensive and reasoning-intensive tasks. We can see that improved training signals, employing different solution strategies for different tasks, potentially different from the one used by the larger model can enhance smaller LMs' reasoning abilities. This research indicates that learning from step-by-step explanations, whether these are generated by humans or more advanced AI models, is a promising direction to improve SLM's capabilities and skills.
    
\subsubsection{\textbf{Explanation Tuning:}} 

Orca1 \footnote{https://arxiv.org/abs/2306.02707} learns from rich explanation traces signal allowing it to overcome the limitations of instruction tuning \footnote{https://arxiv.org/abs/2306.05539}.Once again Orca2 mentioned in the previous section explores how improved training samples can enhance SLMs reasoning ability. These signals are obtained through system instructions crafted to elicit detailed explanations from a teacher model as it reasons through a task. Orca 2 notably outperformed other state-of-the-art models such as LLaMA-2-Chat-13B, WizardLM-13B, and LLaMA-2-70B on various tasks. For instance, Orca-2-13B led LLaMA-2-70B by an average of +3.23 points, which is particularly significant considering Orca 2 has around 5 times fewer parameters. The hallucination rate, evaluated by GPT-4 as the judge, showed that among all versions of Orca 2 and models of comparable size, Orca-2-13B emerged as the most effective model. A lower hallucination rate indicates better performance, reflecting the model's ability to generate more accurate and contextually appropriate responses
    
\subsubsection{\textbf{Knowledge distillation (KD):}} 

KD \cite{hinton2015distilling, Gou_2021} presents an efficient strategy for training neural networks, enabling smaller models to learn complex representations originally captured by larger models. Knowledge distillation involves a two-step training process: First step is, extracting rationales from LLMs: Using techniques like Chain-of-Thought (CoT) prompting to elicit detailed rationales from LLMs, which justify their predicted labels. Secondly, training Smaller Models with Rationales: Utilizing these rationales to train smaller task-specific models, focusing on both label prediction and rationale prediction. Models trained using KD methods outlined in \cite{Zhang2023} and \cite{wang2023improved} outperforms several strong knowledge distillation baselines significantly
    
Symbolic procedural knowledge distillation \cite{li2023symbolic} proved to enhance the implicit knowledge in small language models to facilitate more structured and accurate reasoning. Distilling step-by-step \cite{hsieh2023distilling}, leverages the ability of LLMs to reason about their predictions to train smaller models in a data-efficient way. The author demonstrates that smaller models with 770M-11B parameters can compete with and often surpass the capabilities of larger teacher models in both the original and counterfactual settings. NovaCOMET \cite{west2023novacomet} involves distilling procedural knowledge, which entails decomposing high-level goals into temporally ordered steps, into smaller models using symbolic representations, its performance exceeds comparable open task models like Flan-T5 on a range of commonsense generation tasks. 
    
\subsubsection{\textbf{Reasoning Distillation:}} 

This method is based on "Decompositional distillation" \cite{shridhar2023distilling} training strategy, where a large language model (LLM) decomposes a complex problem into simpler sub-question solutions. On multiple reasoning datasets (GSM8K, StrategyQA, and SVAMP), this distillation strategy boosts the performance of smaller models over 70\% compared to the baselines. Evaluation results on GSM8K dataset demonstrated that all models trained with Decompositional Distillation achieved higher accuracy compared to the Chain of Thought (CoT) baseline. Small model 3B distilled using multi-step reasoning to concentrate their
capacity on a specific target task \cite{fu2023specializing}, outperforms the current 11B and 6B models on the GSM8K test set.
    
\subsubsection{\textbf{Chain-of-Thought Knowledge Distillation:}} 

To teach smaller models to reason\cite{Magister2022TeachingSL}, Chain of Thought (CoT) knowledge distillation technique recommends to perform knowledge distillation by fine-tuning the student model on CoT generated large teacher model and scope the knowledge distillation to a single task due to the limited capacity of the smaller model. Symbolic Chain-of-Thought Distillation (SCoTD) \cite{li2023symbolic}, similar to KD this technique also leverages two step process to first using Chain-of-Thought (CoT) prompting to elicit detailed rationales from LLMs and then used to train small LMs with the rationales. SOCRATIC Chain-Of-Thought (CoT)\cite{shridhar2023distilling} learns a decomposition of the original problem into a sequence of subproblems and uses it to guide the intermediate reasoning steps. SOCRATIC CoT is used to train a combination of two small distilled models: a problem decomposer and a subproblem solver.  SOCRATIC CoT is an effective alternative to CoT, demonstrating cases where a much smaller model (GPT-2 large) can outperform a 10X larger model (GPT-3 6B). SCOTT, or Self-Consistent Chain-of-Thought Distillation \cite{wang2023scott}, is a method for training small language models to generate coherent and consistent rationales. The approach involves using a large teacher model to generate detailed rationales via contrastive decoding, which are then used to train a smaller student model. This method ensures that the student model's predictions are consistent with its own generated rationales.

\subsection{Modularized Training Techniques}

The recent trend in training Large Language Models (LLMs) involving blended ensembles of small LMs and mixture of experts (MoE) models is often referred to as "Modularized Training Techniques". These approaches involve constructing models with different specialized components or 'experts' and blending various models to enhance performance and efficiency. This shift signifies a move towards more flexible, efficient, and task-specific architectures in LLM training, allowing models to address the growing complexity and diversity of language processing tasks more effectively. We discuss about these models in this section

\subsubsection{\textbf{Blended Ensembles:}}

A noticeable trend is employing a combination of smaller models collaboratively to achieve comparable or enhanced performance relative to a single large model. Blended ensembles technique \footnote{https://arxiv.org/abs/2401.02994} involves combining responses from multiple smaller models to create a collective output that rivals or surpasses the performance of much larger models. By blending these models, a more comprehensive and diverse perspective can be obtained, leading to better predictions. Smaller models can be trained to specialize in specific tasks or subsets of the data. By blending these specialized models, a more robust and versatile prediction model can be created. Blended ensembles of 3 models (6-13B) out competes 175B+ ChatGPT model, each smaller model may have different strengths and weaknesses, capturing different aspects of the data. The results from large-scale A/B testing on the Chai research platform revealed that the blended ensemble not only had higher engagement than each of the constituent systems but also outperformed the larger GPT-3.5 model in terms of user engagement and retention. 

\subsubsection{\textbf{Mixture of experts (MoE)}} Mixtral\footnote{https://arxiv.org/abs/2310.06825} employs a novel architecture where each layer consists of multiple feedforward blocks (experts), with a router network selecting two experts per token at each layer. A technique to combine a mixture of expert (MoE) smaller models in Mixtral demonstrate superior capabilities in mathematics, code generation, and tasks that require multilingual understanding, significantly outperforming Llama 2 70B in these domains. The instruct chat model is trained using supervised fine-tuning and Direct Preference optimization, its performance notably surpasses that of GPT-3.5 Turbo, Claude-2.1, Gemini Pro, and Llama 2 70B – chat model on human evaluation benchmarks. Despite its smaller size, Mixtral demonstrates superior performance in mathematics, code generation, and multilingual tasks.

\subsection{Data strategies to train SLMs}

In addition to novel training techniques, we explore data creation strategy and their effective usage in training small language models. 

\subsubsection{\textbf{LLM Generated Synthetic Datasets:}}

TinyStories \footnote{https://arxiv.org/abs/2305.07759}, a synthetic dataset that combines English language elements like grammar, vocabulary, facts, and reasoning used to train small LMs(below 10 million parameters) with simple architectures and yet still produce fluent and consistent english stories. This dataset is used to train very small LMs, or models with minimal transformer layers, to attain factual knowledge and exhibit some extent of reasoning. The dataset's design, which includes generating stories with limited vocabulary and specific features, aims to mimic the language understanding of young children. Zero-Shot Performance Comparison on small LMs, specifically the Orca-2-7B, show either better or comparable performance to larger models like the LLaMA-2-Chat-70B across various reasoning tasks.

Textbooks Are All You Need \footnote{https://arxiv.org/abs/2306.11644}, according to this study on Phi-1,  training data quality plays a critical role in model performance focusing on “textbook-quality” data. The training data mixture contains synthetic datasets specifically created to teach the model common sense reasoning and general knowledge. The synthetic datasets were designed to be diverse and non-repetitive, covering a wide range of coding concepts, skills, scenarios, and varying in difficulty, complexity, and style. This diversity is crucial as it exposes the language model to different ways of expressing and solving problems in code. Despite this small scale, phi-1 attains pass@1 accuracy 50.6\% on HumanEval and 55.5\% on MBPP. This is notable because these scores are competitive with, and in some cases better than, larger models. The phi-1.5 model demonstrates comparable or superior performance to larger models on a range of tasks, For instance, in common sense reasoning benchmarks, phi-1.5 achieves results similar to larger models like Llama2-7B and Vicuna-13B. 

TinyGSM \footnote{https://arxiv.org/abs/2312.09241}, a synthetic dataset of 12.3M grade school math problems paired with Python solutions, generated fully by GPT-3.5. After finetuning on  TinyGSM, 1.3B generation model and a 1.3B verifier model can achieve 81.5 accuracy, outperforming existing models that are orders of magnitude larger. This also rivals the performance of the GPT-3.5 “teacher” model (77.4), from which  model’s training data is generated due to the high-quality dataset TinyGSM and the use of a verifier model, which selects the final outputs from multiple candidate generations.

\subsubsection{\textbf{Common Crawl Internet Datasets:}}

Pile \cite{gao2020pile}, 825 GiB english corpus dataset created using common crawl technique from sources like PubMed Central, ArXiv, GitHub,
the FreeLaw Project, Stack Exchange, the US Patent etc is used to train SLMs like Cerebras-GPT \footnote{https://arxiv.org/abs/2304.03208} family of models (111M to 13B parameters). The Pile dataset's comprehensive and diverse nature has been instrumental in achieving these results, providing a robust training ground for the Cerebras-GPT models across various scales. We conclude that model selection for synthetic data generation, prompt engineering techniques in data generation, diversity and robustness of the samples, selective filtering in curating the datasets have significant impact on the quality and performance of SLMs.

\subsection{Post-Training Optimizations for SLM Development}

 This subsection delves into post-training optimizations and pivotal techniques like quantization in creating small yet effective language models. We provide an overview of  post-training optimizations which have emerged as effective strategies for reducing model size while preserving or even improving their performance.

\subsubsection{\textbf{Quantization:}} 

Quantization is reducing the precision of the numbers used to represent a model's weights, activations, or both, this reduction in numerical precision helps decrease the model's size, memory footprint and computational requirements, enabling faster processing and reduced storage. SmoothQuant \cite{xiao2023smoothquant}, a training-free, accuracy-preserving, and general-purpose post-training quantization (PTQ) solution to enable 8-bit weight, 8-bit activation (W8A8) quantization for LLMs. SmoothQuant enables an INT8 quantization of both weights and activations for all the matrix multiplications, demonstrates up to 1.56x speedup and 2x memory reduction with negligible loss in accuracy. GPTQ \cite{frantar2023gptq}, a new one-shot weight quantization method based on approximate second-order information, highly-accurate and highly-efficient method that provides quantization to 2-bit or even ternary quantization levels. Activation-aware Weight Quantization (AWQ) \cite{lin2023awq}, approach for low-bit weight-only quantization for the optimal per-channel scaling that protects the salient weights by observing the activation, not weights.

\subsubsection{\textbf{Model Pruning:}} 

Model pruning is a critical technique for reducing the size of language models, making them more suitable for deployment in resource-constrained environments. Bi-level Optimization (BIP) \cite{Zhang2023} effectively reduces the size of language models by achieving up to 74\% sparsity in models like ResNet-20, ResNet-56, and ResNet-18, while maintaining or even improving accuracy. Notably, BIP finds the best winning ticket in nearly all settings and is up to 5 times faster than Iterative Magnitude Pruning (IMP), demonstrating both its efficiency in model size reduction and its computation speed advantage. Human-in-the-loop(HIL) model pruning method \cite{10248433}, which improves the sparse training effect of the model by introducing Gaussian penalty terms, and combines automatic and manual method to set pruning threshold to meet the goal of model lightweighting. Based on the pruned and quantized model, the size of the model is reduced by 87.94\% compared to the original model (YOLOv5), with only a 1.48\% decrease in accuracy. After deployment optimization on GPUs, the real-time performance is improved by 2.28 times. Sheared-llama is another example of a pruned model where the authors use two key techniques: (1) targeted structured pruning, which reduces a larger model to a specified target shape by eliminating layers, heads, and intermediate and hidden dimensions in an end-to-end manner, and (2) dynamic batch loading, which entails updating the composition of sampled data in each training batch dynamically, based on varying losses across different domains. They showcase the effectiveness of their approach through the development of the Sheared-LLaMA series, which includes pruning the LLaMA2-7B model down to 1.3B and 2.7B parameters. These Sheared-LLaMA models surpass the performance of equivalent-sized state-of-the-art open-source models such as Pythia, INCITE, and OpenLLaMA in various downstream and instruction tuning evaluations, while requiring only 3\% of the compute resources needed for training such models from scratch \cite{xia2023sheared}.

\subsection{SLMs as Draft models}
Draft models have emerged as a critical component in the evolution of language model inference, particularly in the context of speculative decoding. These models serve as lightweight alternatives to larger language models, designed to accelerate the inference process while maintaining acceptable levels of output quality. In the realm of small language models (SLMs), draft models represent a particularly compelling approach, as they leverage the efficiency advantages of reduced parameter counts while still contributing to sophisticated language generation tasks.\cite{draftm} 

The fundamental principle behind draft models lies in their ability to generate preliminary sequences of tokens that can be subsequently verified by larger models. This two-stage process, comprising drafting and verification phases, allows for significant optimization of computational resources. Draft models typically operate with substantially fewer parameters than their larger counterparts, enabling them to generate token sequences more rapidly. This efficiency stems from reduced memory access requirements and simplified architectural designs, making them particularly suitable for resource-constrained environments. For example in Leviathan \emph{et. al.}, we see how inference on a 11B T5 model can be accelerated to 3.4x the baseline with speculative decoding. \cite{leviathan2023fast} 

In the context of small language models, draft models can be categorized into two primary architectural approaches: independent and dependent drafters. Independent draft models function as standalone entities, often implemented as smaller versions of existing model architectures. For instance, a common approach involves utilizing a reduced-scale version from the same model family, such as employing OPT-125M as a draft model for OPT-7B. This approach benefits from architectural similarity and shared pretraining experiences, leading to better alignment in prediction behaviors. Recent innovations in independent draft models include the development of specialized architectures like Chimera, which combines trigram encoders for short-range dependencies with full context encoders for managing longer-range relationships.

Dependent draft models, conversely, integrate directly with the primary model architecture. This integration can take various forms, such as the addition of specialized drafting heads or the implementation of early exit mechanisms. Notable examples include the Medusa architecture, which incorporates multiple decoding heads to generate subsequent tokens in parallel, and Hydra, which extends this concept by enabling each head to consider previously speculated tokens within the continuation. These approaches eliminate the need for separate model maintenance while potentially offering more efficient resource utilization.

The technical implementation of draft models faces several key challenges that current research aims to address. One primary concern is the trade-off between drafting speed and accuracy, often measured through the Effective Decoding Length (EDL). Higher EDL values indicate more successful draft sequences but must be balanced against computational overhead. Another significant challenge lies in maintaining alignment between draft and target models, particularly in independent drafting approaches. Methods such as knowledge distillation and online adaptation have been proposed to enhance this alignment, though perfect synchronization remains an open challenge. 

Recent advancements in draft model techniques have introduced several innovative approaches to improve efficiency and effectiveness. These include the development of dynamic adaptation mechanisms that adjust draft generation based on context and computational resources, the implementation of specialized training objectives to optimize draft quality, and the exploration of hybrid architectures that combine multiple drafting strategies. For instance, S2D (Sorted Speculative Decoding) employs adaptive sub-model selection based on specific tasks, while online speculative decoding continuously adjusts the draft model based on the query distribution. Several other options for speculative decoding have appeared in the recent past, including lookahead decoding, Medusa, and streaming speculative decoding with the same fundamental technical principal. \cite{spec1, spec2, spec3, spec4, spec5, spec6, spec7}

Looking ahead, the field of draft models faces several important challenges that warrant further research attention. These include improving generalization across different tasks, reducing the computational overhead of verification processes, and developing more efficient methods for handling long-context scenarios. Additionally, the integration of draft models with emerging hardware architectures and the optimization of their performance in resource-constrained environments remain active areas of investigation. As language models continue to evolve, the role of draft models in enabling efficient inference while maintaining output quality will likely become increasingly significant, particularly in the context of small language models where resource efficiency is paramount. We refer the readers interested in more details to the following survey papers dedicated to this topic of speculative decoding and draft models\footnote{https://arxiv.org/abs/2411.13157}. \cite{specsurvey}

\subsection{SLMs for agentic workflows}

 Agentic AI systems impose unique requirements that distinguish them from traditional conversational AI: precise tool
  calling for external system integration, structured output generation for downstream processing, efficient reasoning for complex
  multi-step planning, behavioral alignment for consistent formatting, and cost-effective deployment for scaled
  operations. Unlike broad-purpose language models, agentic systems decompose complex tasks into specialized,
  repetitive subtasks that can be handled more efficiently by smaller, focused models. This fundamental shift has positioned Small Language
  Models (SLMs) as a compelling alternative to Large Language Models (LLMs) for agentic applications, offering superior cost-efficiency,
  lower latency, and enhanced deployability while meeting the specific functional requirements of agent-based
  workflows. This paradigm shift has positioned Small Language Models (SLMs) as a compelling alternative to Large Language Models (LLMs) for many agentic applications, offering superior cost-efficiency, lower latency, and enhanced deployability while maintaining adequate task performance. This section examines how SLMs address each of these critical agentic requirements,
  demonstrating their suitability for powering the next generation of AI agent systems.

 The core argument for SLMs in agentic systems rests on three fundamental principles - 1. they are sufficiently powerful to handle the narrow language modeling tasks required by agents, 2. inherently more operationally suitable for repetitive and constrained workflows, and 3. necessarily more economical due to their reduced computational requirements.

Models we have covered earlier are already suited well to agentic AI tasks. For example, the Microsoft Phi series represents a breakthrough in compact model design for agentic applications. Phi-2 (2.7B parameters) achieves commonsense reasoning and code generation performance comparable to 30B parameter models while delivering approximately 15× faster inference speeds~\footnote{https://huggingface.co/microsoft/phi-2}. Phi-3 Small (7B parameters) extends these capabilities, matching language understanding and reasoning performance of models up to 70B parameters from the same generation~\footnote{https://arxiv.org/abs/2404.14219}. These models excel in tool calling and instruction following—critical capabilities for agentic workflows that require precise communication across model-tool interfaces. The NVIDIA Nemotron-H series (2B, 4.8B, and 9B parameters) employs hybrid Mamba-Transformer architectures specifically optimized for agentic tasks, achieving instruction following and code generation accuracy comparable to dense 30B LLMs while requiring an order-of-magnitude fewer inference FLOPs~\footnote{https://www.nvidia.com/en-us/ai-data-science/foundation-models/nemotron/}. The hybrid architecture provides particular advantages in handling sequential decision-making tasks common in agentic systems.

Specialized tool-calling models demonstrate the potential for SLMs to exceed LLM performance in focused domains. Salesforce xLAM-2-8B achieves state-of-the-art tool calling performance despite its modest 8B parameter size, surpassing frontier models like GPT-4o and Claude 3.5 in specialized benchmarks~\cite{zhang2024xlam}. NVIDIA Hymba-1.5B, utilizing Mamba-attention hybrid architectures, outperforms 13B models in instruction following while providing 3.5× greater token throughput~\cite{dong2024hymba}. The DeepSeek-R1-Distill series (1.5B-8B parameters) represents a new generation of reasoning-focused SLMs trained on outputs from larger reasoning models, with the 7B variant notably outperforming proprietary models like Claude-3.5-Sonnet and GPT-4o on reasoning benchmarks~\footnote{https://arxiv.org/abs/2501.12948}. These examples demonstrate how knowledge distillation and architectural innovations can transfer complex reasoning capabilities to smaller models suitable for agentic deployment.

The economic advantages of SLMs in agentic systems are substantial, providing substantial cost reduction compared to 70-175B LLMs in inference latency, energy consumption, and computational requirements. This efficiency enables real-time agentic responses at scale, particularly crucial for applications requiring rapid tool invocation and decision-making cycles. Fine-tuning agility represents another key advantage, as parameter-efficient techniques like LoRA allow SLM behaviors to be added, fixed, or specialized overnight rather than over weeks~\cite{hu2021lora}. The compact nature of SLMs enables edge deployment scenarios using frameworks like ChatRTX, allowing for offline agentic inference with enhanced data privacy and reduced latency. Additionally, SLMs demonstrate more efficient parameter utilization, with a larger proportion of their parameters actively contributing to inference compared to LLMs, which often exhibit sparse activation patterns~\cite{song2024powerinfer}.

Agentic systems require strict adherence to formatting requirements for tool calls and structured outputs, making behavioral alignment a critical consideration. SLMs can be fine-tuned with specific formatting constraints at low cost, ensuring consistent behavioral alignment that is often more challenging to achieve with general-purpose LLMs. This is particularly important because agentic interactions frequently involve parsing by downstream code components that expect specific formats, making hallucinatory formatting mistakes potentially catastrophic. The narrow functionality exposed by agents—essentially heavily instructed gateways to language models—means that a SLM appropriately fine-tuned for selected prompts can suffice while providing the benefits of increased efficiency and greater flexibility.

Modern agentic architectures increasingly employ heterogeneous model compositions, where SLMs handle routine tasks while LLMs are reserved for complex reasoning. This modular approach, supported by frameworks that enable multiple model endpoints, optimizes both cost and performance across diverse agentic workflows. The natural heterogeneity of agentic systems allows for different models to be invoked at different stages, with language models themselves serving as tools that can be called by other language models. Agentic systems also provide natural pathways for gathering high-quality training data through operational logs, which can be systematically collected, filtered, and used to fine-tune specialized SLMs, creating a continuous improvement cycle that reduces dependency on larger models over time~\cite{schick2023toolformer}.

Despite current limitations in open-domain reasoning and complex multi-step planning scenarios, recent advances in inference-time scaling, tool augmentation, and structured reasoning approaches suggest these gaps are rapidly narrowing~\cite{zhou2022least}. The flexibility of SLMs enables rapid iteration and adaptation to evolving user needs, supporting new behaviors, meeting new output formatting requirements, and complying with changing regulatory requirements in different markets. This democratization effect allows more individuals and organizations to participate in developing specialized language models for agentic systems, potentially reducing systemic biases and encouraging innovation through increased competition~\cite{jungherr2023artificial}. As training techniques continue advancing and deployment infrastructure evolves toward more modular, specialized systems that favor SLMs' strengths while mitigating their limitations through careful task decomposition, SLMs are increasingly well-suited for many agentic AI applications, particularly 
those involving repetitive, well-scoped subtasks where cost efficiency and latency 
are prioritized. However, complex multi-step planning and open-domain reasoning 
remain areas where larger models retain significant advantages, and the extent to 
which SLMs can close these gaps through task decomposition and tool augmentation 
remains an active area of research.

\section{Discussion}
 
As we have seen in the sections above, general-purpose or task-agnostic SLMs are performing at levels that are similar to  LLMs that are over 10 times larger. However papers rarely go into the discussion of why this might be the case, with the exception of one clear insight from the paper discussed in our data strategy section above. 
 
\subsection{Caveats on Benchmark Comparisons}
\label{sec:caveats}
 
Before interpreting the performance comparisons presented in this section, several important limitations of current evaluation practices should be noted. First, \textbf{benchmark contamination} is a well-documented concern: models trained on web-scale corpora likely encounter benchmark data during pre-training, leading to inflated scores that do not reflect genuine generalization~\cite{sainz2023contamination}. Many of the models surveyed here are trained on datasets derived from Common Crawl and similar web sources that may overlap with benchmarks such as MMLU and GSM8K. Second, \textbf{metric choice can create misleading comparisons} across model scales. Schaeffer~et~al.~\cite{schaeffer2023mirage} showed that apparent emergent abilities can be artifacts of nonlinear metrics (such as exact-match accuracy) rather than genuine capability transitions. Since many benchmarks cited in this survey rely on exact-match or pass/fail metrics, performance gaps between SLMs and LLMs may appear more dramatic or more narrow than underlying capability differences warrant. Third, \textbf{GSM8K results should be interpreted with caution}. Mirzadeh~et~al.~\cite{mirzadeh2024gsmsymbolic} showed that LLM performance on GSM8K drops significantly when only numerical values are changed, and can collapse by up to 65\% when a single irrelevant clause is added, suggesting that strong scores may partially reflect pattern matching rather than robust reasoning. Where we state that a smaller model ``outperforms'' a larger one, this claim is scoped to the specific benchmarks and evaluation settings reported in the cited paper.
 
\subsection{Effective Size Analysis}
 
We analyze the \emph{effective size} of these SLMs based on the performance claims presented in the papers discussed above. To calculate the "effective size" of the various models listed in the table, we utilized available data such as reported model sizes, performance claims, and comparisons with known benchmarks. Details of how we interpret effective size can be found in the appendix. 
 
\begin{figure}[h!]
 
\includegraphics[width=\linewidth]{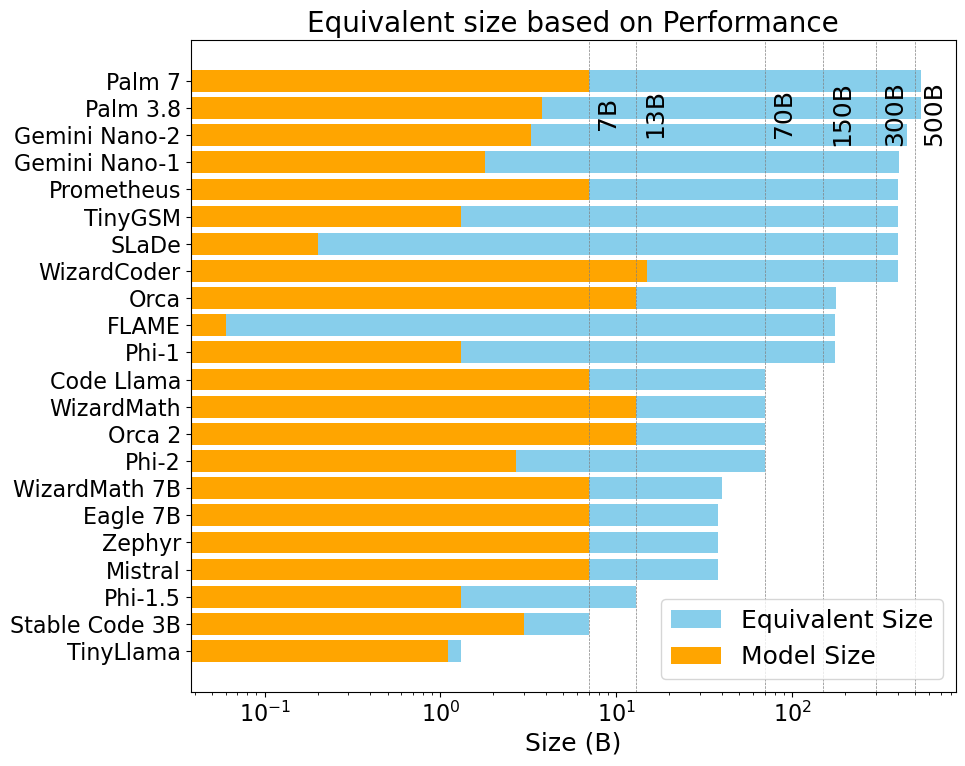}
\vspace{-0.7cm}
\caption{Equivalent sizes of SLMs based on performance benchmarks; more details in Table \ref{tab:model-specs}}
\vspace{-0.1cm}
\label{fig:eqsize}
\end{figure}
 
The Palm (540B) serves as a reference point, and we anchor effective size calculations to it as follows~\cite{chowdhery2023palm}. For models like Palm 2 (3.8B, 7B), where it's stated that they outperform Palm, we denote their equivalent size as 540B, implying that despite their smaller parameter count, they achieve comparable or superior performance. Similarly, when comparing models across different architectures, adjustments are made based on performance differentials. For instance, GPT-4, which outperforms Palm 2, is scaled by the maximum performance difference observed (GSM8K), leading to an estimated effective size of around 640B. Furthermore, models like GPT-3.5, ChatGPT, and others mentioned in papers (such as GPT4.5) lack publicly known sizes, hence their effective sizes are inferred based on reported performance relative to known models. By considering these factors and interpolating between known anchor points, we approximate the effective sizes of the models listed in the Table \ref{tab:model-specs}, providing a comparative measure of their capabilities despite variations in reported parameter counts. It's essential to note that while some models' performance claims are provided in papers, their exact sizes remain undisclosed, requiring us to rely on performance benchmarks and comparisons to estimate their effective sizes accurately. While Table \ref{tab:model-specs} provides more details on benchmarks used and claims, here we depict actual model sizes and estimated effective model sizes on a log scale as a rough illustrative comparison. We caveat the observations in figure \ref{fig:eqsize} with the following: 1. when calculating effective size, we use the best case scenario, that is, we capture the best performing benchmark as reported in individual papers; in general, performance across other benchmarks may be lower (or significantly lower) compared to the same base models; 2. Model sizes that are not publicly available are interpolated once again using performance on best reported benchmarks in individual papers.
 
We emphasize that these effective size estimates have significant methodological limitations. A model that outperforms a larger model on a narrow, task-specific benchmark (e.g., FLAME outperforming Codex on formula tasks) does not possess the broad capabilities of that larger model. Furthermore, the factors that enable SLMs to close performance gaps with LLMs---including training data quality, fine-tuning regime, quantization, and architectural innovations, each represent independent variables that are collapsed into a single comparison in Table~\ref{tab:model-specs}. We present these estimates as an intuitive visualization of SLM competitiveness rather than a rigorous quantitative framework.
 
Across several papers, we observe that higher quality data that is either human curated or LLM generated can be used to effectively train better performing SLMs. Quoting the Palm paper "PaLM 2-L, is significantly smaller than the largest PaLM model but uses more training compute" based on scaling laws \cite{chowdhery2023palm}. While some papers and articles claim that SLMs \textit{break} existing scaling laws from Kaplan and Chinchilla, we maintain that the laws need to be revised to capture the actual capabilities of models across parameter size, dataset size and quality. For example , the Chinchilla scaling law could be modified to $L(N, D, Q) = E + AN^\alpha + B(DQ)^\beta
$ where $Q$ represents the quality of the dataset, introducing a more nuanced view of how data attributes beyond quantity can impact model performance. What remains to be answered is if there is a way to objectively assess quality $Q$, but it is possible to use this as an additional tuneable parameter to satisfy \textit{surprising} SLM performance. This suggestion is not entirely novel, as well have seen the scaling laws being modified to adjust to other cost functions, for example inference throughput \cite{sardana2023beyond}.  
 
From section \ref{sec:tsslm} we see that models that are task-specific still overshadow larger models such as WizardMath models (7B) outperforming Llama 2 70B in mathematical reasoning tasks, or CodeLlama (7B) outperforming Llama 2 70B on coding tasks. The predominant motivation for SLM development remains to be their use in real-world applications where computational resources are limited, such as mobile devices, edge computing, and in regions with limited internet connectivity. For example, SmolLM represents a groundbreaking family of small language models available in three sizes (135M, 360M, and 1.7B parameters), designed for efficient local device operation while maintaining impressive performance. SmolLM-360M outperforms all sub-500M parameter models, and SmolLM-1.7B leads among models under 2B parameters, including strong Python coding capabilities. The models' efficient design allows them to run on devices with limited memory, making them suitable for smartphones and laptops, while instruction-tuned versions demonstrate competitive performance.
 
Inference on alternative hardware and CPUs are still being explored, but are promising. For example, by taking advantage extreme quantization and sparse training, popular models like Llama can be used for inference on CPUs \cite{pmlr-v119-kurtz20a}. While LLM use in training and inference is still widespread, SLMs are continuing to demonstrate outperforming much larger counterparts, and their  price-to-performance during training as well as inference is attractive. 
 
\subsection{Limitations of This Survey and Open Challenges}
\label{sec:limitations}
 
This survey has several limitations that readers should consider. Although we surveyed approximately 160 papers, our selection was not conducted under a formal systematic review protocol (e.g., PRISMA), and the rapid pace of SLM research means that notable models released after early 2026 may be absent. In particular, the OLMo model family~\cite{olmo2024} from AI2, which emphasizes fully open training data and reproducibility, is not covered. Many performance claims originate from technical reports published by the organizations that developed the models, and we have not independently verified these results. As discussed in Section~\ref{sec:caveats}, benchmark contamination, metric sensitivity, and the fragility of mathematical reasoning scores all affect the reliability of comparisons presented here. Additionally, most benchmarks evaluate English-language performance; the multilingual capabilities of SLMs remain insufficiently evaluated. This survey also does not include a systematic evaluation of frontier LLMs (e.g., GPT-4o, Claude 3.5 Sonnet, Llama 3.1 405B) on the same benchmarks under controlled conditions, so claims that SLMs ``rival'' larger models are based on cross-paper comparisons rather than head-to-head evaluation. Open challenges where SLMs have not yet demonstrated competitive performance include long-context reasoning beyond 32K tokens, complex multi-step planning, and robustness to adversarial inputs. Safety evaluation of SLMs, including hallucination rates, bias, and harmful content generation, also remains underdeveloped.
 
Looking ahead, several promising avenues exist for SLM research. A key area is leveraging SLMs' unique properties and constraints during training, building on successes like knowledge distillation and low-bit quantization to develop customized approaches that can capture reasoning capabilities of larger models in compact SLM architectures. Training separate capabilities, and then merging SLMs also seems promising (see Mergekit here - \url{https://github.com/arcee-ai/mergekit}). Crucially, SLMs must be evaluated across a wide range of benchmarks beyond language tasks, like multimodal understanding, safety considerations, and highly specialized tasks where SOTA LLMs struggle, such as using LLMs as judges. From a practical view, optimizing SLM deployment on specialized hardware like mobile and neuromorphic chips could enable new edge and IoT use cases. Moreover, investigating computationally efficient continual updating of SLMs could allow their use in dynamic environments with evolving data/knowledge, where it is impractical to use LLMs.
\vspace{-0.1cm}

\begin{table*}[h!]
\caption{Equivalent model sizes based on performance benchmarks}
\centering
\small  
\begin{tabular}{|p{2cm}|p{1cm}|p{4.8cm}|p{1.7cm}|p{5cm}|}
\hline
\textbf{Model} & \textbf{Size (B)} & \textbf{Performance Claim} & \textbf{Equivalent Size (B)} & \textbf{Eval Datasets} \\ \hline
Llama2 & 7 & Used as a baseline & 7 & Math, MMLU, BBH, AGI Eval, TriviaQA, Natural Questions, Big Bench hard, Human Eval, GSM8k \\ \hline
Tiny Llama & 1.1 & Performance around other 1.3 B models & 1.3 & HellaSwag, OpenBookQA, ARC, BoolQ, PIQA, MMLU \\ \hline
Mistral & 7 & Effective Llama size is around 38B & 38 & MMLU, HellaSwag, WinoG, PIQA, Arc-e, Arc-c, NQ, TriviaQA, HumanEval, MBPP, MATH, GSM8K \\ \hline
Zephyr & 7 & Outperforms Mistral & 38 & MT-bench, Alpaca eval \\ \hline
Phi-1 & 1.3 & Outperforms Codex-12B, CodeGen-Mono-16.1B, PaLM-Coder-540B, GPT-3.5 175B & 175 & HumanEval, MBPP \\ \hline
Phi-1.5 & 1.3 & Comparable to Llama2 7B, Falcon 7B, Vicuna 13B & 13 & WinoGrande, ARC-Easy, ARC-Challenge, BoolQ, SIQA \\ \hline
Phi-2 & 2.7 & Close to Llama2 70B & 70 & Common sense reasoning, language understanding, math, coding \\ \hline
Orca & 13 & Up to 88\% of ChatGPT & 176 & AGIeval, BigBench, BBH, SAT, LSAT, GRE, GMAT \\ \hline
Orca 2 & 13 & Outperforms WizardLM 70B, Llama2 chat 70B & 70 & Reasoning, math, knowledge understanding, safety, truthfulness \\ \hline
Gemini Nano-1 & 1.8 & Surpasses USM, Whisper (except FLEURS) and close to Gemini Pro & 405 & 50 benchmarks across capabilities: Factuality, Long-Context, Math/Science, Reasoning, Multilingual \\ \hline
Gemini Nano-2 & 3.25 & Strong performance on factuality and reasoning, STEM, coding, multimodal and multilingual tasks & 450 & 50 benchmarks across capabilities including BoolQ, Natural Questions, Big Bench, MMLU \\ \hline
MoE Mistral & 47 & Outperforms Llama2 70B, GPT-3.5 & 400 & MMLU, Hellaswag, ARC, Winogrande, MBPP, GSM8K, MT Bench \\ \hline
Eagle 7B & 7 & Outperforms Mistral 7B & 38 & Cross-lingual benchmarks \\ \hline
WizardMath 7B & 7 & Effective size up to 40B, surpasses most 7B-40B open-source models & 40 & GSM8K, MATH \\ \hline
WizardMath 13B & 13 & Surpasses Llama2 70B & 70 & GSM8K, MATH \\ \hline
WizardCoder & 15 & Superior to Anthropic's Claude, Google's Bard & 400 & HumanEval, HumanEval+, MBPP, DS-1000 \\ \hline
Code Llama & 7 & Outperforms Llama2 70B & 70 & HumanEval, MBPP \\ \hline
Stable Code 3B & 3 & On par with Code Llama 7B & 7 & Multi-PL \\ \hline
SLaDe & 0.2 & Outperforms Ghidra, ChatGPT, BTC & 400 & ExeBench, AnghaBench decompilation \\ \hline
ALMA-13B-R & 13 & Surpasses GPT-4, WMT winners & 640 & Neural machine translation \\ \hline
FLAME & 0.06 & Outperforms Codex models in 6/10 settings & 175 & Formula repair, auto-completion, syntax \\ \hline
BioGPT & 0.335 & Outperforms few-shot GPT-4 & 640 & PubMedQA \\ \hline
ChatLaw & 13 & Outperforms GPT-4, Lawyer LLaMA & 640 & Chinese legal examples \\ \hline
TinyGSM & 1.3 & Rivals GPT-3.5 teacher model & 400 & GSM8K \\ \hline
Prometheus & 7 & Performance comparable to GPT3.5 with human evaluation & 400 & Feedback bench, Vivuna bench, MT bench, Flask Eval \\ \hline
Palm 3.8 & 3.8 & Beats the original Palm 1 on some tasks & 540 & TriviaQA, Natural Questions, Hellaswag, LAMBADA, StoryCloze, etc. \\ \hline
Palm 7 & 7 & Beats the original Palm 1 & 540 & TriviaQA, Natural Questions, Hellaswag, LAMBADA, StoryCloze, etc. \\ \hline
\end{tabular}
\label{tab:model-specs}
\end{table*}

\section{Other Related Work}
\label{app:2}

While the focus of this survey is SLMs, related methods like Parameter Efficient Fine Tuning (PEFT), adapters, and mixture of experts are gaining traction due to similar benefits; they use fewer resources for training and inference, even with comparable total size to LLMs. PEFT methods add or reparameterize the base model to fine-tune performance. One popular PEFT model is Low-Rank Adaptation (LoRA) \cite{hu2021lora}, where trainable rank decomposition matrices are injected into each Transformer layer, reducing active parameters and memory footprint. S-LoRA \footnote{https://arxiv.org/abs/2311.03285} is a scalable LoRA serving system enabling thousands of LoRA adapters on single/multi-GPUs.
Other PEFT methods include MultiLoRA \footnote{https://arxiv.org/abs/2311.11501}, SAID \cite{Aghajanyan2020IntrinsicDE},  prompt tuning \cite{Li2021PrefixTuningOC}, LeTS \cite{Fu2021LearntoShareAH}, and adapter architectures like AdapterHub \footnote{https://github.com/a2aproject/A2A}. Hybrid methods combine multiple PEFT approaches, e.g., MAM, Compactor\cite{Davison2021CompacterEL,he2021towards}.
Weight-Decomposed Low-Rank Adaptation (DoRA) \cite{liu2024dora} closes the gap between full fine-tuning and LoRA by decomposing pre-trained weights into magnitude and direction components. AdaMix \footnote{https://arxiv.org/abs/2205.12410} takes inspiration from mixture of experts models to introduce multiple PEFT modules per Transformer layer, routing inputs stochastically.
Alternative architectures like Mamba \footnote{https://arxiv.org/abs/2312.00752} use structured state space sequence models without attention or MLP blocks, outperforming transformers on various benchmarks. BitNet b1.58 \footnote{https://arxiv.org/abs/2402.17764} is a 1.58-bit SLM variant with ternary parameter values, offering superior latency, memory efficiency, throughput, and energy savings over FP16 LLMs. OpenELM \footnote{https://arxiv.org/abs/2404.14619} leverages layer-wise scaling and instruction tuning to outperform existing open LLMs while using fewer pre-training tokens. LM-Guided CoT \cite{lee2024small} uses a lightweight LM to guide a larger LM for reasoning tasks, improving performance through reinforcement learning and knowledge distillation.

\section{Conclusion}
This survey has demonstrated that Small Language Models (SLMs) are increasingly proving their capability to match or exceed the performance of much larger models across various tasks and domains. Through innovative training techniques like knowledge distillation, progressive learning, and explanation tuning, combined with effective data strategies and post-training optimizations, SLMs are achieving remarkable efficiency while maintaining high performance levels.

Several key insights emerge from our analysis. First, the quality of training data appears to be as crucial as, if not more important than, quantity - as evidenced by models like Phi-1 and TinyStories achieving strong results with carefully curated datasets. Second, task-specific SLMs consistently demonstrate the ability to outperform larger general-purpose models in their specialized domains, suggesting a promising direction for practical applications. Third, our illustrative analysis of effective model sizes suggests that some SLMs can approach performance levels associated with 
significantly larger models on specific benchmarks, though these 
comparisons are limited by best-case benchmark selection and should 
not be interpreted as general capability equivalence. This 
nonetheless suggests the need for revised scaling frameworks that 
account for factors like data quality.

\clearpage
\appendix
\section{Appendix}
\label{app:1}
\begin{center}

\begin{table}[h!]
\centering
\caption{Comparison of Hybrid State Space Models and Efficient Architectures}
\centering
\begin{tabular}{|l|p{2.5cm}p{2.5cm}p{2.5cm}p{2.5cm}|}
\hline
\textbf{Aspect} & \textbf{Hymba} & \textbf{Zamba} & \textbf{Jamba} & \textbf{Mamba} \\
\hline
Fusion & Parallel fusion of & Sequential with & Interleaved transformer & Pure selective \\
Approach & attention and SSM & shared global & and Mamba layers & SSM without \\
& within each layer & attention every N layers & with MoE & attention \\
\hline
Efficiency & Meta-tokens and & Parameter sharing & MoE for sparse & Hardware-aware \\
Optimizations & cross-layer KV & in global & parameter & parallel scan \\
& cache sharing & attention & activation & algorithm \\
\hline
Performance- & Balanced with & Lower memory & Sparse activation & Linear complexity \\
Memory & meta-tokens and & from shared & via MoE & from pure SSM \\
Tradeoff & KV sharing & attention & & \\
\hline
Model Scale/ & 1.5B & 7B & 7B base with & Up to 3B \\
Complexity & parameters & parameters & 52B total via MoE & parameters tested \\
\hline
\end{tabular}
\label{table:model_comparison}
\end{table}
\end{center}
\begin{table*}[h!]

\caption{Sample completions from small models with the input ``The diagonals of a rectangle..."}

\vskip 0.15in
\begin{center}
\begin{small}
\begin{tabular}{|p{1.5cm}|p{2cm}p{8cm}p{3cm}|}
\hline
\textbf{Model} & \textbf{Size} & \textbf{Completion} & \textbf{Reference} \\
\hline
TinyStories    & 33M  & \textbf{\textit{The diagonals of a rectangle}} and they were so beautiful   &  \footnote{https://arxiv.org/abs/2305.07759}            \\
distilGPT2     & 88M   & \textbf{\textit{The diagonals of a rectangle}} As a rule, there's no rule to call "corridal" when an extra &      \cite{sanh2019distilbert}              \\
GPT Neo & 125M &
  \textbf{\textit{The diagonals of a rectangle}}  are represented by a set of diagonals, each of which is represented by a set of diagonals. & \cite{gao2020pile}
   \\
GPT2 &
  137M &
  \textbf{\textit{The diagonals of a rectangle}}  of three is the square-joint which you wish to point toward the face of the object. & OpenAI Blog
   \\
Bloom &
  560M &
  \textbf{\textit{The diagonals of a rectangle}}  are the points of the rectangle that are the intersection of the diagonal with the line segment connecting the points. & BigScience Workshop 2022
   \\
TinyLlama      & 1.1B                       & \textbf{\textit{The diagonals of a rectangle}}  are the lines that intersect the rectangle at right angles. &   Tiny LLaMA                 \\
Phi 1.5      & 1.3B                       & \textbf{\textit{The diagonals of a rectangle}}  are the lines that connect opposite corners. &   Microsoft Research Technical Report                 \\
GPT Neo        & 1.37B                      & \textbf{\textit{The diagonals of a rectangle}}  or diagonals, are the two opposite sides.                   &       \cite{gao2020pile}             \\

Stable LM      & 1.6B                      & \textbf{\textit{The diagonals of a rectangle}}  are congruent. If the length of the rectangle is 16 cm, what is the perimeter of the rectangle, in centimeters?                &       StableLM technical report             \\

Phi 2      & 2B                       & \textbf{\textit{The diagonals of a rectangle}}  are the line segments that connect opposite corners of the rectangle. &   Microsoft Research Technical Report                 \\
GPT Neo        & 2.7 B                      & \textbf{\textit{The diagonals of a rectangle}} are drawn in two shades of blue or green.                   &         \cite{gao2020pile}           \\
Stable LM      & 3B                      & \textbf{\textit{The diagonals of a rectangle}}  are drawn from one corner of the rectangle to the opposite corner, creating a right angle at each endpoint.                &       StableLM technical report            \\
Mistral        & 7B                         & \textbf{\textit{The diagonals of a rectangle}} are equal.                                                 & \footnote{https://arxiv.org/abs/2310.06825}\\       
Zephyr        & 7B                         & \textbf{\textit{The diagonals of a rectangle}} are lines that connect opposite vertices of a rectangle, forming two diagonal lines that bisect the rectangle.                                                 & \cite{tunstall2023zephyr}\\   

\hline
\end{tabular}
\end{small}
\end{center}
\vskip -0.1in
\label{tab:1}
\end{table*}

\clearpage
\appendix
\label{app:2}

\begin{figure*}[h!]
    \resizebox{0.75\textwidth}{!}{
        \begin{forest}
forked edges,
for tree={
  grow=east,
  folder,
  reversed=true,
  edge path={
    \noexpand\path [draw, \forestoption{edge}]
    (!u.parent anchor) -- +(5pt,0) |- (.child anchor)\forestoption{edge label};
  },
  anchor=base west,
  parent anchor=east,
  child anchor=west,
  base=center,
  font=\large,
  rectangle,
  draw=hidden-draw,
  rounded corners,
  align=center,
  text centered,
  minimum width=5em,
  edge+={darkgray, line width=1pt},
  s sep=3pt,
  inner xsep=2pt,
  inner ysep=3pt,
  line width=0.8pt,
  fork sep=0.15cm,
  tier/.option=level,
},
where level=1{text width=30em,font=\normalsize, fill=,}{},
where level=2{text width=15em,font=\normalsize, fill=white,}{},
where level=3{text width=35em,font=\normalsize, fill=white,}{},
[
  {Eval Datasets}, root style
  [
    {General Knowledge \& Reasoning}, child style
    [
      {MMLU}, grandchild style
      [
        {Llama2}, leaf style
      ]
      [
        {Tiny Llama}, leaf style
      ]
      [
        {Mistral}, leaf style
      ]
      [
        {Phi-1.5}, leaf style
      ]
      [
        {Gemini Nano-1}, leaf style
      ]
      [
        {Gemini Nano-2}, leaf style
      ]
      [
        {Palm 3.8}, leaf style
      ]
      [
        {Palm 7}, leaf style
      ]
    ]
    [
      {Big Bench Hard (BBH)}, grandchild style
      [
        {Llama2}, leaf style
      ]
      [
        {Orca}, leaf style
      ]
    ]
    [
      {Natural Questions}, grandchild style
      [
        {Llama2}, leaf style
      ]
      [
        {Mistral}, leaf style
      ]
      [
        {Gemini Nano-1}, leaf style
      ]
      [
        {Gemini Nano-2}, leaf style
      ]
      [
        {Palm 3.8}, leaf style
      ]
      [
        {Palm 7}, leaf style
      ]
    ]
    [
      {TriviaQA}, grandchild style
      [
        {Llama2}, leaf style
      ]
      [
        {Mistral}, leaf style
      ]
      [
        {Palm 3.8}, leaf style
      ]
      [
        {Palm 7}, leaf style
      ]
    ]
    [
      {PIQA}, grandchild style
      [
        {Tiny Llama}, leaf style
      ]
    ]
    [
      {BoolQ}, grandchild style
      [
        {Tiny Llama}, leaf style
      ]
      [
        {Gemini Nano-1}, leaf style
      ]
      [
        {Gemini Nano-2}, leaf style
      ]
    ]
    [
      {HellaSwag}, grandchild style
      [
        {Tiny Llama}, leaf style
      ]
      [
        {Mistral}, leaf style
      ]
    ]
    [
      {OpenBookQA}, grandchild style
      [
        {Tiny Llama}, leaf style
      ]
    ]
    [
      {ARC}, grandchild style
      [
        {Tiny Llama}, leaf style
      ]
      [
        {Phi-1.5}, leaf style
      ]
    ]
  ]
  [
    {Math \& STEM}, child style
    [
      {MATH}, grandchild style
      [
        {Llama2}, leaf style
      ]
      [
        {Mistral}, leaf style
      ]
      [
        {WizardMath 7B}, leaf style
      ]
      [
        {WizardMath 13B}, leaf style
      ]
    ]
    [
      {GSM8K}, grandchild style
      [
        {Llama2}, leaf style
      ]
      [
        {Mistral}, leaf style
      ]
      [
        {Phi-2}, leaf style
      ]
      [
        {WizardMath 7B}, leaf style
      ]
      [
        {WizardMath 13B}, leaf style
      ]
      [
        {TinyGSM}, leaf style
      ]
    ]
    [
      {SciBench}, grandchild style
      [
        {Gemini Nano-1}, leaf style
      ]
      [
        {Gemini Nano-2}, leaf style
      ]
    ]
  ]
  [
    {Coding \& Programming}, child style
    [
      {HumanEval}, grandchild style
      [
        {Llama2}, leaf style
      ]
      [
        {Mistral}, leaf style
      ]
      [
        {Phi-1}, leaf style
      ]
      [
        {Code Llama}, leaf style
      ]
      [
        {WizardCoder}, leaf style
      ]
    ]
    [
      {MBPP}, grandchild style
      [
        {Mistral}, leaf style
      ]
      [
        {Phi-1}, leaf style
      ]
      [
        {Code Llama}, leaf style
      ]
    ]
    [
      {DS-1000}, grandchild style
      [
        {WizardCoder}, leaf style
      ]
    ]
    [
      {Multi-PL}, grandchild style
      [
        {Stable Code 3B}, leaf style
      ]
    ]
  ]
  [
    {Language Understanding \& Common Sense}, child style
    [
      {WinoGrande}, grandchild style
      [
        {Mistral}, leaf style
      ]
      [
        {Phi-1.5}, leaf style
      ]
    ]
    [
      {ARC-Challenge}, grandchild style
      [
        {Phi-1.5}, leaf style
      ]
    ]
    [
      {ARC-Easy}, grandchild style
      [
        {Phi-1.5}, leaf style
      ]
    ]
    [
      {SIQA}, grandchild style
      [
        {Phi-1.5}, leaf style
      ]
    ]
    [
      {LAMBADA}, grandchild style
      [
        {Palm 3.8}, leaf style
      ]
      [
        {Palm 7}, leaf style
      ]
    ]
    [
      {StoryCloze}, grandchild style
      [
        {Palm 3.8}, leaf style
      ]
      [
        {Palm 7}, leaf style
      ]
    ]
  ]
  [
    {Legal \& Professional}, child style
    [
      {AGI Eval (legal tasks)}, grandchild style
      [
        {Llama2}, leaf style
      ]
      [
        {Orca}, leaf style
      ]
      [
        {ChatLaw}, leaf style
      ]
    ]
  ]
  [
    {Factuality \& Retrieval}, child style
    [
      {BoolQ}, grandchild style
      [
        {Tiny Llama}, leaf style
      ]
      [
        {Gemini Nano-1}, leaf style
      ]
      [
        {Gemini Nano-2}, leaf style
      ]
    ]
    [
      {Natural Questions}, grandchild style
      [
        {Llama2}, leaf style
      ]
      [
        {Mistral}, leaf style
      ]
      [
        {Gemini Nano-1}, leaf style
      ]
      [
        {Gemini Nano-2}, leaf style
      ]
      [
        {Palm 3.8}, leaf style
      ]
      [
        {Palm 7}, leaf style
      ]
    ]
    [
      {TriviaQA}, grandchild style
      [
        {Llama2}, leaf style
      ]
      [
        {Mistral}, leaf style
      ]
      [
        {Palm 3.8}, leaf style
      ]
      [
        {Palm 7}, leaf style
      ]
    ]
    [
      {Big Bench}, grandchild style
      [
        {Gemini Nano-1}, leaf style
      ]
      [
        {Gemini Nano-2}, leaf style
      ]
    ]
  ]
]
\end{forest}
}
    \caption{SLM Eval dataset tree.}
    \label{fig:keplm_taxonomy}
\end{figure*}

\forestset{
    root style/.style={draw, thick, rounded corners, fill=gray!30},
    child style/.style={draw, rounded corners, fill=blue!20},
    grandchild style/.style={draw, fill=green!20},
    greatgrandchild style/.style={draw, fill=yellow!20},
    leaf style/.style={draw, fill=red!20},
}
\tikzset{
  my-box/.style={
    rectangle, draw=gray!60, rounded corners, minimum height=1.5em,
    minimum width=40em, inner sep=2pt, align=center, line width=0.8pt,
  },
  leaf/.style={
    my-box, minimum height=1.5em, text=black, align=center,
    font=\normalsize, inner xsep=2pt, inner ysep=4pt, line width=0.8pt,
  }
}

\begin{figure*}[h!]
\centering
\resizebox{0.95\textwidth}{!}{%
\begin{forest}
forked edges,
for tree={
  grow=east,
  folder,
  reversed=true,
  edge path={
    \noexpand\path [draw, \forestoption{edge}]
    (!u.parent anchor) -- +(5pt,0) |- (.child anchor)\forestoption{edge label};
  },
  anchor=base west,
  parent anchor=east,
  child anchor=west,
  base=center,
  font=\large,
  rectangle,
  draw=gray!40,
  rounded corners,
  align=center,
  text centered,
  minimum width=5em,
  edge+={darkgray, line width=1pt},
  s sep=3pt,
  inner xsep=2pt,
  inner ysep=3pt,
  line width=0.8pt,
  fork sep=0.15cm,
  tier/.option=level,
},
where level=1{text width=15em,font=\normalsize, fill=gray!15,}{},
where level=2{text width=15em,font=\normalsize, fill=blue!10,}{},
where level=3{text width=22em,font=\normalsize, fill=green!10,}{},
where level=4{text width=55em,font=\small, fill=yellow!10,}{}
[{SLM Model Performance}, root style
  [{Size: 0.5B}, child style
    [{Model Type: Base}, grandchild style
      [{Qwen1.5-0.5B}, greatgrandchild style
        [{AVERAGE: 5.35\% | IFEVAL: 17.06\% | BBH: 5.04\% | MATH: 1.74\% | GPQA: 0.56\% | MUSR: 4.30\% | MMLU: 3.41\%}, leaf style]
      ]
      [{Qwen2-0.5B}, greatgrandchild style
        [{AVERAGE: 7.22\% | IFEVAL: 18.73\% | BBH: 7.92\% | MATH: 2.64\% | GPQA: 1.45\% | MUSR: 4.60\% | MMLU: 8.00\%}, leaf style]
      ]
      [{Qwen2.5-0.5B}, greatgrandchild style
        [{AVERAGE: 6.55\% | IFEVAL: 16.27\% | BBH: 6.95\% | MATH: 3.93\% | GPQA: 0.00\% | MUSR: 2.08\% | MMLU: 10.06\%}, leaf style]
      ]
    ]
    [{Model Type: Instruct}, grandchild style
      [{Qwen1.5-0.5B-Chat}, greatgrandchild style
        [{AVERAGE: 5.68\% | IFEVAL: 18.07\% | BBH: 4.32\% | MATH: 0.68\% | GPQA: 2.57\% | MUSR: 6.06\% | MMLU: 2.36\%}, leaf style]
      ]
      [{Qwen2-0.5B-Instruct}, greatgrandchild style
        [{AVERAGE: 6.59\% | IFEVAL: 22.47\% | BBH: 5.88\% | MATH: 2.87\% | GPQA: 0.00\% | MUSR: 2.41\% | MMLU: 5.90\%}, leaf style]
      ]
      [{Qwen2.5-0.5B-Instruct}, greatgrandchild style
        [{AVERAGE: 10.11\% | IFEVAL: 31.53\% | BBH: 8.17\% | MATH: 10.35\% | GPQA: 1.23\% | MUSR: 1.37\% | MMLU: 8.00\%}, leaf style]
      ]
    ]
  ]
  [{Size: 1B}, child style
    [{Model Type: Base}, grandchild style
      [{Llama-3.2-1B}, greatgrandchild style
        [{AVERAGE: 4.20\% | IFEVAL: 14.78\% | BBH: 4.37\% | MATH: 1.21\% | GPQA: 0.00\% | MUSR: 2.56\% | MMLU: 2.26\%}, leaf style]
      ]
    ]
    [{Model Type: Instruct}, grandchild style
      [{Llama-3.2-1B-Instruct}, greatgrandchild style
        [{AVERAGE: 14.44\% | IFEVAL: 56.98\% | BBH: 8.74\% | MATH: 7.02\% | GPQA: 3.36\% | MUSR: 2.97\% | MMLU: 7.58\%}, leaf style]
      ]
    ]
  ]
  [{Size: 1.3B}, child style
    [{Model Type: Base}, grandchild style
      [{phi-1}, greatgrandchild style
        [{AVERAGE: 5.57\% | IFEVAL: 20.68\% | BBH: 4.27\% | MATH: 0.98\% | GPQA: 2.01\% | MUSR: 3.70\% | MMLU: 1.80\%}, leaf style]
      ]
      [{phi-1\_5}, greatgrandchild style
        [{AVERAGE: 7.17\% | IFEVAL: 20.33\% | BBH: 7.47\% | MATH: 1.81\% | GPQA: 2.35\% | MUSR: 3.39\% | MMLU: 7.68\%}, leaf style]
      ]
    ]
  ]
  [{Size: 1.5B}, child style
    [{Model Type: Base}, grandchild style
      [{DeepSeek-R1-Distill-Qwen-1.5B}, greatgrandchild style
        [{AVERAGE: 10.35\% | IFEVAL: 34.63\% | BBH: 4.73\% | MATH: 16.92\% | GPQA: 0.78\% | MUSR: 2.97\% | MMLU: 2.08\%}, leaf style]
      ]
      [{Qwen2-1.5B}, greatgrandchild style
        [{AVERAGE: 10.45\% | IFEVAL: 21.13\% | BBH: 11.78\% | MATH: 7.02\% | GPQA: 1.90\% | MUSR: 3.59\% | MMLU: 17.24\%}, leaf style]
      ]
      [{Qwen2.5-1.5B}, greatgrandchild style
        [{AVERAGE: 13.85\% | IFEVAL: 26.74\% | BBH: 16.66\% | MATH: 9.14\% | GPQA: 4.70\% | MUSR: 5.27\% | MMLU: 20.61\%}, leaf style]
      ]
    ]
    [{Model Type: Instruct}, grandchild style
      [{Qwen2-1.5B-Instruct}, greatgrandchild style
        [{AVERAGE: 14.14\% | IFEVAL: 33.71\% | BBH: 13.70\% | MATH: 7.18\% | GPQA: 1.57\% | MUSR: 12.03\% | MMLU: 16.68\%}, leaf style]
      ]
      [{Qwen2.5-1.5B-Instruct}, greatgrandchild style
        [{AVERAGE: 18.43\% | IFEVAL: 44.76\% | BBH: 19.81\% | MATH: 22.05\% | GPQA: 0.78\% | MUSR: 3.19\% | MMLU: 19.99\%}, leaf style]
      ]
    ]
  ]
  [{Size: 1.8B}, child style
    [{Model Type: Base}, grandchild style
      [{Qwen1.5-1.8B}, greatgrandchild style
        [{AVERAGE: 9.27\% | IFEVAL: 21.54\% | BBH: 9.76\% | MATH: 3.17\% | GPQA: 7.38\% | MUSR: 3.96\% | MMLU: 9.80\%}, leaf style]
      ]
    ]
    [{Model Type: Instruct}, grandchild style
      [{Qwen1.5-1.8B-Chat}, greatgrandchild style
        [{AVERAGE: 9.26\% | IFEVAL: 20.19\% | BBH: 5.91\% | MATH: 1.96\% | GPQA: 6.38\% | MUSR: 12.18\% | MMLU: 8.93\%}, leaf style]
      ]
    ]
  ]
  [{Size: 2B}, child style
    [{Model Type: Base}, grandchild style
      [{codegemma-1.1-2b}, greatgrandchild style
        [{AVERAGE: 7.13\% | IFEVAL: 22.94\% | BBH: 7.55\% | MATH: 1.28\% | GPQA: 2.01\% | MUSR: 5.93\% | MMLU: 3.09\%}, leaf style]
      ]
      [{gemma-2-2b}, greatgrandchild style
        [{AVERAGE: 10.13\% | IFEVAL: 19.93\% | BBH: 11.76\% | MATH: 2.87\% | GPQA: 1.68\% | MUSR: 11.43\% | MMLU: 13.11\%}, leaf style]
      ]
    ]
    [{Model Type: Instruct}, grandchild style
      [{gemma-1.1-2b-it}, greatgrandchild style
        [{AVERAGE: 8.05\% | IFEVAL: 30.67\% | BBH: 5.86\% | MATH: 1.81\% | GPQA: 2.57\% | MUSR: 2.02\% | MMLU: 5.37\%}, leaf style]
      ]
      [{gemma-2-2b-it}, greatgrandchild style
        [{AVERAGE: 17.05\% | IFEVAL: 56.68\% | BBH: 17.98\% | MATH: 0.08\% | GPQA: 3.24\% | MUSR: 7.08\% | MMLU: 17.22\%}, leaf style]
      ]
      [{gemma-2b-it}, greatgrandchild style
        [{AVERAGE: 7.49\% | IFEVAL: 26.90\% | BBH: 5.21\% | MATH: 2.04\% | GPQA: 3.80\% | MUSR: 3.03\% | MMLU: 3.92\%}, leaf style]
      ]
    ]
  ]
  [{Size: 2.7B}, child style
    [{Model Type: Base}, grandchild style
      [{phi-2}, greatgrandchild style
        [{AVERAGE: 15.53\% | IFEVAL: 27.39\% | BBH: 28.04\% | MATH: 2.95\% | GPQA: 2.91\% | MUSR: 13.84\% | MMLU: 18.09\%}, leaf style]
      ]
    ]
  ]
  [{Size: 3B}, child style
    [{Model Type: Base}, grandchild style
      [{Llama-3.2-3B}, greatgrandchild style
        [{AVERAGE: 8.70\% | IFEVAL: 13.37\% | BBH: 14.23\% | MATH: 1.89\% | GPQA: 2.35\% | MUSR: 3.81\% | MMLU: 16.53\%}, leaf style]
      ]
      [{Qwen2.5-3B}, greatgrandchild style
        [{AVERAGE: 18.10\% | IFEVAL: 26.90\% | BBH: 24.30\% | MATH: 14.80\% | GPQA: 6.38\% | MUSR: 11.76\% | MMLU: 24.48\%}, leaf style]
      ]
    ]
    [{Model Type: Instruct}, grandchild style
      [{Llama-3.2-3B-Instruct}, greatgrandchild style
        [{AVERAGE: 24.20\% | IFEVAL: 73.93\% | BBH: 24.06\% | MATH: 17.67\% | GPQA: 3.80\% | MUSR: 1.37\% | MMLU: 24.39\%}, leaf style]
      ]
      [{Qwen2.5-3B-Instruct}, greatgrandchild style
        [{AVERAGE: 27.16\% | IFEVAL: 64.75\% | BBH: 25.80\% | MATH: 36.78\% | GPQA: 3.02\% | MUSR: 7.57\% | MMLU: 25.05\%}, leaf style]
      ]
    ]
  ]
  [{Size: 3.8B}, child style
    [{Model Type: Instruct}, grandchild style
      [{Phi-3-mini-128k-instruct}, greatgrandchild style
        [{AVERAGE: 26.34\% | IFEVAL: 59.76\% | BBH: 37.10\% | MATH: 14.05\% | GPQA: 9.06\% | MUSR: 7.71\% | MMLU: 30.38\%}, leaf style]
      ]
      [{Phi-3-mini-4k-instruct}, greatgrandchild style
        [{AVERAGE: 25.97\% | IFEVAL: 56.13\% | BBH: 39.27\% | MATH: 11.63\% | GPQA: 9.28\% | MUSR: 7.64\% | MMLU: 31.85\%}, leaf style]
      ]
      [{Phi-3.5-mini-instruct}, greatgrandchild style
        [{AVERAGE: 28.18\% | IFEVAL: 57.75\% | BBH: 36.75\% | MATH: 19.64\% | GPQA: 11.97\% | MUSR: 10.10\% | MMLU: 32.91\%}, leaf style]
      ]
      [{Phi-4-mini-instruct}, greatgrandchild style
        [{AVERAGE: 29.41\% | IFEVAL: 73.78\% | BBH: 38.74\% | MATH: 16.99\% | GPQA: 7.94\% | MUSR: 6.45\% | MMLU: 32.58\%}, leaf style]
      ]
    ]
  ]
  [{Size: 4B}, child style
    [{Model Type: Base}, grandchild style
      [{Qwen1.5-4B}, greatgrandchild style
        [{AVERAGE: 11.77\% | IFEVAL: 24.45\% | BBH: 16.25\% | MATH: 5.29\% | GPQA: 3.58\% | MUSR: 4.82\% | MMLU: 16.22\%}, leaf style]
      ]
    ]
    [{Model Type: Instruct}, grandchild style
      [{Qwen1.5-4B-Chat}, greatgrandchild style
        [{AVERAGE: 12.63\% | IFEVAL: 31.57\% | BBH: 16.30\% | MATH: 2.79\% | GPQA: 2.24\% | MUSR: 7.36\% | MMLU: 15.51\%}, leaf style]
      ]
    ]
  ]
  [{Size: 7B}, child style
    [{Model Type: Base}, grandchild style
      [{DeepSeek-R1-Distill-Qwen-7B}, greatgrandchild style
        [{AVERAGE: 14.99\% | IFEVAL: 40.38\% | BBH: 7.88\% | MATH: 19.56\% | GPQA: 3.91\% | MUSR: 3.55\% | MMLU: 14.68\%}, leaf style]
      ]
      [{Llama-2-7b-hf}, greatgrandchild style
        [{AVERAGE: 8.81\% | IFEVAL: 25.19\% | BBH: 10.35\% | MATH: 1.74\% | GPQA: 2.24\% | MUSR: 3.76\% | MMLU: 9.57\%}, leaf style]
      ]
      [{Mistral-v0.1-7B}, greatgrandchild style
        [{AVERAGE: 14.58\% | IFEVAL: 23.86\% | BBH: 22.02\% | MATH: 2.95\% | GPQA: 5.59\% | MUSR: 10.68\% | MMLU: 22.36\%}, leaf style]
      ]
      [{Mistral-v0.3-7B}, greatgrandchild style
        [{AVERAGE: 14.23\% | IFEVAL: 22.66\% | BBH: 24.04\% | MATH: 3.02\% | GPQA: 5.59\% | MUSR: 8.36\% | MMLU: 21.70\%}, leaf style]
      ]
      [{Orca-2-7b}, greatgrandchild style
        [{AVERAGE: 14.40\% | IFEVAL: 21.83\% | BBH: 22.43\% | MATH: 1.96\% | GPQA: 1.45\% | MUSR: 24.09\% | MMLU: 14.65\%}, leaf style]
      ]
      [{Qwen1.5-7B}, greatgrandchild style
        [{AVERAGE: 16.02\% | IFEVAL: 26.84\% | BBH: 23.08\% | MATH: 9.29\% | GPQA: 6.49\% | MUSR: 9.16\% | MMLU: 21.29\%}, leaf style]
      ]
      [{Qwen2-7B}, greatgrandchild style
        [{AVERAGE: 23.93\% | IFEVAL: 31.49\% | BBH: 34.71\% | MATH: 20.39\% | GPQA: 7.27\% | MUSR: 14.32\% | MMLU: 35.37\%}, leaf style]
      ]
      [{gemma-7b}, greatgrandchild style
        [{AVERAGE: 15.44\% | IFEVAL: 26.59\% | BBH: 21.12\% | MATH: 7.40\% | GPQA: 4.92\% | MUSR: 10.98\% | MMLU: 21.64\%}, leaf style]
      ]
    ]
    [{Model Type: Instruct}, grandchild style
      [{Llama-2-7b-chat-hf}, greatgrandchild style
        [{AVERAGE: 9.61\% | IFEVAL: 39.86\% | BBH: 4.46\% | MATH: 1.96\% | GPQA: 0.45\% | MUSR: 3.28\% | MMLU: 7.64\%}, leaf style]
      ]
      [{Mistral-v0.1-7B-Instruct}, greatgrandchild style
        [{AVERAGE: 12.77\% | IFEVAL: 44.87\% | BBH: 7.65\% | MATH: 2.27\% | GPQA: 0.00\% | MUSR: 6.13\% | MMLU: 15.72\%}, leaf style]
      ]
      [{Mistral-v0.2-7B-Instruct}, greatgrandchild style
        [{AVERAGE: 18.51\% | IFEVAL: 54.96\% | BBH: 22.91\% | MATH: 3.02\% | GPQA: 3.47\% | MUSR: 7.61\% | MMLU: 19.08\%}, leaf style]
      ]
      [{Mistral-v0.3-7B-Instruct}, greatgrandchild style
        [{AVERAGE: 19.23\% | IFEVAL: 54.65\% | BBH: 25.57\% | MATH: 3.85\% | GPQA: 3.91\% | MUSR: 4.30\% | MMLU: 23.06\%}, leaf style]
      ]
      [{Phi-3-small-128k-instruct}, greatgrandchild style
        [{AVERAGE: 31.97\% | IFEVAL: 63.68\% | BBH: 45.63\% | MATH: 20.26\% | GPQA: 8.95\% | MUSR: 14.50\% | MMLU: 38.78\%}, leaf style]
      ]
      [{Phi-3-small-8k-instruct}, greatgrandchild style
        [{AVERAGE: 32.34\% | IFEVAL: 64.97\% | BBH: 46.21\% | MATH: 18.87\% | GPQA: 8.28\% | MUSR: 16.77\% | MMLU: 38.96\%}, leaf style]
      ]
      [{Qwen1.5-7B-Chat}, greatgrandchild style
        [{AVERAGE: 17.62\% | IFEVAL: 43.71\% | BBH: 22.38\% | MATH: 6.27\% | GPQA: 7.05\% | MUSR: 4.64\% | MMLU: 21.68\%}, leaf style]
      ]
      [{Qwen2-7B-Instruct}, greatgrandchild style
        [{AVERAGE: 27.94\% | IFEVAL: 56.79\% | BBH: 37.81\% | MATH: 27.64\% | GPQA: 6.38\% | MUSR: 7.37\% | MMLU: 31.64\%}, leaf style]
      ]
      [{gemma-1.1-7b-it}, greatgrandchild style
        [{AVERAGE: 17.69\% | IFEVAL: 50.39\% | BBH: 15.93\% | MATH: 4.91\% | GPQA: 5.82\% | MUSR: 11.51\% | MMLU: 17.60\%}, leaf style]
      ]
      [{gemma-7b-it}, greatgrandchild style
        [{AVERAGE: 13.07\% | IFEVAL: 38.68\% | BBH: 11.94\% | MATH: 2.95\% | GPQA: 4.59\% | MUSR: 12.53\% | MMLU: 7.72\%}, leaf style]
      ]
    ]
  ]
  [{Size: 8B}, child style
    [{Model Type: Base}, grandchild style
      [{Llama-3-8B}, greatgrandchild style
        [{AVERAGE: 13.63\% | IFEVAL: 14.55\% | BBH: 24.50\% | MATH: 4.53\% | GPQA: 7.38\% | MUSR: 6.24\% | MMLU: 24.55\%}, leaf style]
      ]
      [{Llama-3.1-8B}, greatgrandchild style
        [{AVERAGE: 14.42\% | IFEVAL: 12.46\% | BBH: 25.30\% | MATH: 6.57\% | GPQA: 8.05\% | MUSR: 8.72\% | MMLU: 25.42\%}, leaf style]
      ]
    ]
    [{Model Type: Instruct}, grandchild style
      [{Llama-3-8B-Instruct}, greatgrandchild style
        [{AVERAGE: 23.91\% | IFEVAL: 74.08\% | BBH: 28.24\% | MATH: 8.69\% | GPQA: 1.23\% | MUSR: 1.60\% | MMLU: 29.60\%}, leaf style]
      ]
      [{Llama-3.1-8B-Instruct}, greatgrandchild style
        [{AVERAGE: 23.76\% | IFEVAL: 49.22\% | BBH: 29.38\% | MATH: 15.56\% | GPQA: 8.72\% | MUSR: 8.61\% | MMLU: 31.09\%}, leaf style]
      ]
      [{Ministral-8B-Instruct-2410}, greatgrandchild style
        [{AVERAGE: 24.19\% | IFEVAL: 58.96\% | BBH: 25.82\% | MATH: 19.56\% | GPQA: 4.59\% | MUSR: 10.72\% | MMLU: 25.46\%}, leaf style]
      ]
    ]
  ]
  [{Size: 9B}, child style
    [{Model Type: Base}, grandchild style
      [{gemma-2-9b}, greatgrandchild style
        [{AVERAGE: 21.21\% | IFEVAL: 20.40\% | BBH: 34.10\% | MATH: 13.44\% | GPQA: 10.51\% | MUSR: 14.30\% | MMLU: 34.48\%}, leaf style]
      ]
    ]
    [{Model Type: Instruct}, grandchild style
      [{gemma-2-9b-it}, greatgrandchild style
        [{AVERAGE: 32.07\% | IFEVAL: 74.36\% | BBH: 42.14\% | MATH: 19.49\% | GPQA: 14.77\% | MUSR: 9.74\% | MMLU: 31.95\%}, leaf style]
      ]
    ]
  ]
]
\end{forest}%
}
\end{figure*}

\begin{figure}[h!]
\includegraphics[width=\linewidth, trim=0 0 0 1cm, clip]{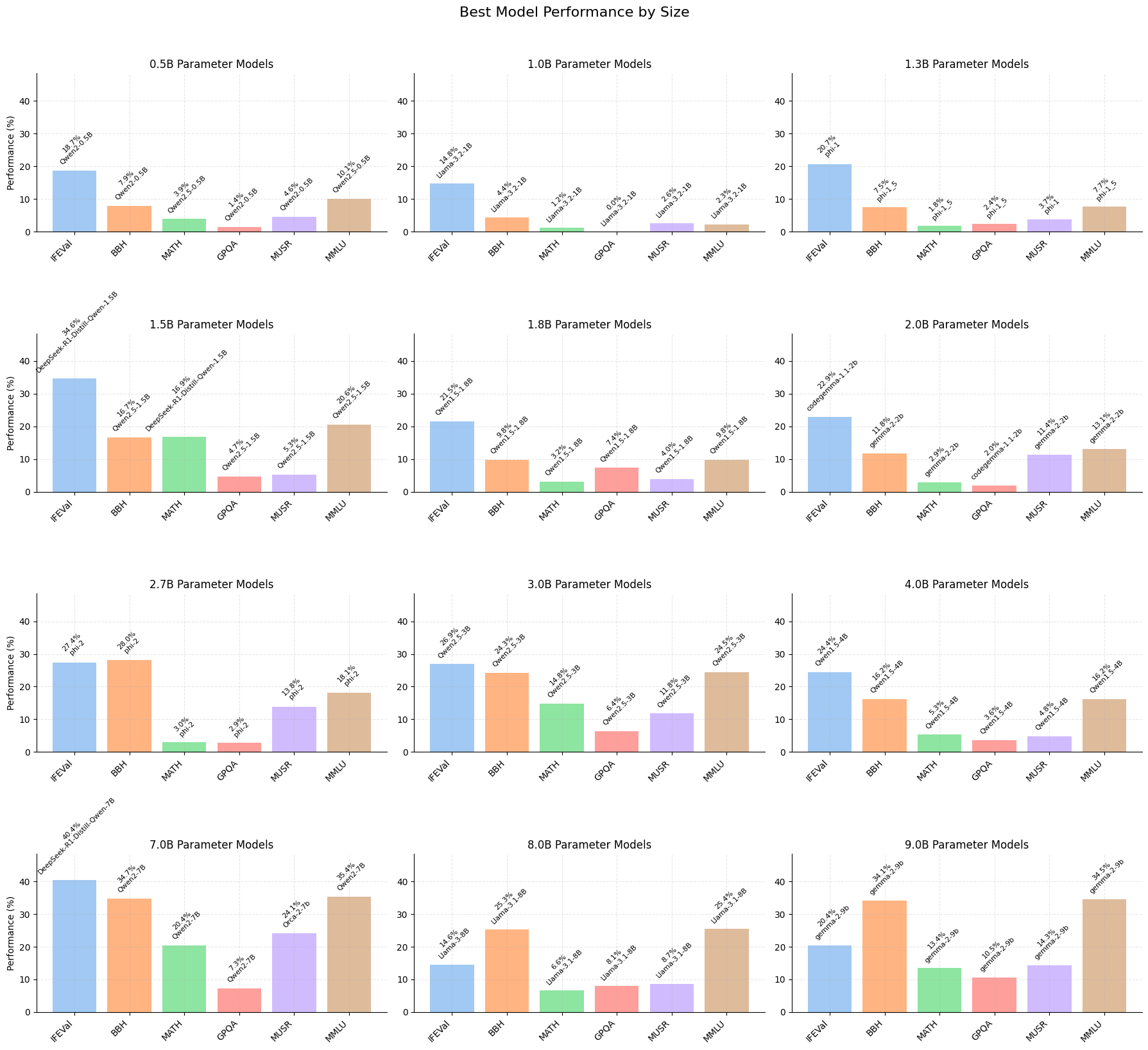}
\vspace{-0.7cm}
\caption{Best base model by size}
\vspace{-0.1cm}
\label{fig:basemodel}
\end{figure}

\end{document}